\documentclass[lettersize,journal]{IEEEtran}
\usepackage{amsmath,amsfonts,stmaryrd,amssymb}
\usepackage{algorithmic}
\usepackage{array}
\usepackage[caption=false,font=normalsize,labelfont=sf,textfont=sf]{subfig}
\usepackage{textcomp}
\usepackage{stfloats}
\usepackage{url}
\usepackage{verbatim}
\usepackage{graphicx}
\def\BibTeX{{\rm B\kern-.05em{\sc i\kern-.025em b}\kern-.08em
    T\kern-.1667em\lower.7ex\hbox{E}\kern-.125emX}}
\usepackage{balance}
\usepackage{cite}

\usepackage[nameinlink,capitalize]{cleveref}
\usepackage[table]{xcolor}
\usepackage{bmpsize}
\usepackage{amsfonts,amsmath,stmaryrd,amssymb}
\usepackage{booktabs}
\usepackage{dsfont}
\usepackage{mathrsfs}
\usepackage{multirow}
\usepackage{tikz}
\usepackage{stmaryrd}
\usepackage{upgreek}
\usepackage{enumitem}
\usepackage{graphicx}
\usepackage{multicol}
\usepackage{ntheorem} 
\usepackage{lscape} 
\usepackage{xspace}
\usepackage{listings}
\usepackage{etoc} % for table of content for appendix only
\usepackage[ruled,vlined]{algorithm2e}
\SetKwRepeat{Do}{do}{while}
\usepackage{subfig}
\usepackage[nameinlink,capitalize]{cleveref}
\usepackage{bbm}
\usepackage{tabularx}
\usepackage{makecell}

\usepackage[final]{pdfpages}

\usepackage{mathtools}

\usepackage{xspace}
\makeatletter

\DeclareRobustCommand\onedot{\futurelet\@let@token\bmv@onedotaux}
\def\bmv@onedotaux{\ifx\@let@token.\else.\null\fi\xspace}

\def\eg{\emph{e.g}\onedot} 
\def\ie{\emph{i.e}\onedot} 

\makeatother

\newcommand{\defeq}{\vcentcolon=}

\renewcommand{\Pr}{\mathbb{P}}
\newcommand{\E}[2]{{\mathbb{E}_{#1}}\left[ #2 \right]}

\newcommand{\HSf}{\mathcal{H}}

\newcommand{\LCal}{\mathscr{L}}

\newcommand{\Real}{\mathbb{R}}

\renewcommand{\d}{\mathrm{d}}

\newcommand{\MBPcomp}{\Xi}

\newcommand{\HPObserVolume}{\mathsf{C}}

\newcommand{\obsTime}{o}

\newcommand{\ppllt}{{\textsc{PP-LL}}}
\newcommand{\npllt}{{\textsc{IC-LL}}}

\newcommand{\ppll}{\LCal_{\ppllt}}
\newcommand{\npll}{\LCal_{\npllt}}

\newcommand{\pmbp}{\textsc{PCMHP}\xspace}

\newcommand{\ignore}[1]{}

\newcommand{\actived}{\texttt{ACTIVE}\xspace}

\makeatletter
\newcommand{\xMapsto}[2][]{\ext@arrow 0599{\Mapstofill@}{#1}{#2}}
\def\Mapstofill@{\arrowfill@{\Mapstochar\Relbar}\Relbar\Rightarrow}
\makeatother

\newcommand{\alexs}[1]{#1}
\newcommand{\pioc}[1]{#1}

\newcommand{\BlackBox}{\rule{1.5ex}{1.5ex}}  % end of proof

\newtheorem{theorem}{Theorem}

\newtheorem{definition}[theorem]{Definition}

\begin{document}
\title{Linking Across Data Granularity: Fitting Multivariate Hawkes Processes to Partially Interval-Censored Data}
\author{Pio Calderon, Alexander Soen, Marian-Andrei Rizoiu}
\author{Pio Calderon, Alexander Soen, Marian-Andrei Rizoiu
\thanks{Pio Calderon and Marian-Andrei Rizoiu are with the University of Technology Sydney}
\thanks{Alexander Soen is with the Australian National University.}
}

\markboth{IEEE Transactions on Computational Social Systems,~Vol.~18, No.~9, September~2020}%
{How to Use the IEEEtran \LaTeX \ Templates}

\maketitle

\begin{abstract}
  The multivariate Hawkes process (MHP) is widely used for analyzing data streams that interact with each other, where events generate new events within their own dimension (via self-excitation) or across different dimensions (via cross-excitation). However, in certain applications, the timestamps of individual events in some dimensions are unobservable, and only event counts within intervals are known, referred to as partially interval-censored data. 
  The MHP is unsuitable for handling such data since its estimation requires event timestamps. 
  In this study, we introduce the Partially Censored Multivariate Hawkes Process (\pmbp), a novel point process which shares parameter equivalence with the MHP and can effectively model both timestamped and interval-censored data. 
  We demonstrate the capabilities of the \pmbp using synthetic and real-world datasets. 
  Firstly, we illustrate that the \pmbp can approximate MHP parameters
  and recover the spectral radius 
  using synthetic event histories. 
  Next, we assess the performance of the \pmbp in predicting YouTube popularity and find that \pioc{the \pmbp outperforms the popularity estimation algorithm Hawkes Intensity Process (HIP) \cite{Rizoiu2017}. 
  Comparing with the fully interval-censored HIP, we show that the \pmbp improves prediction performance by accounting for point process dimensions, particularly when there exist significant cross-dimension interactions.} 
  Lastly, we leverage the \pmbp to gain qualitative insights from a dataset comprising daily COVID-19 case counts from multiple countries and COVID-19-related news articles. 
  By clustering the \pmbp-modeled countries, we unveil hidden interaction patterns between occurrences of COVID-19 cases and news reporting.
\end{abstract}


% !TEX root=../main.tex
%
\section{Introduction}
\label{section:introduction}

\begin{figure}[t]
    \centering
    \includegraphics[width=\columnwidth]{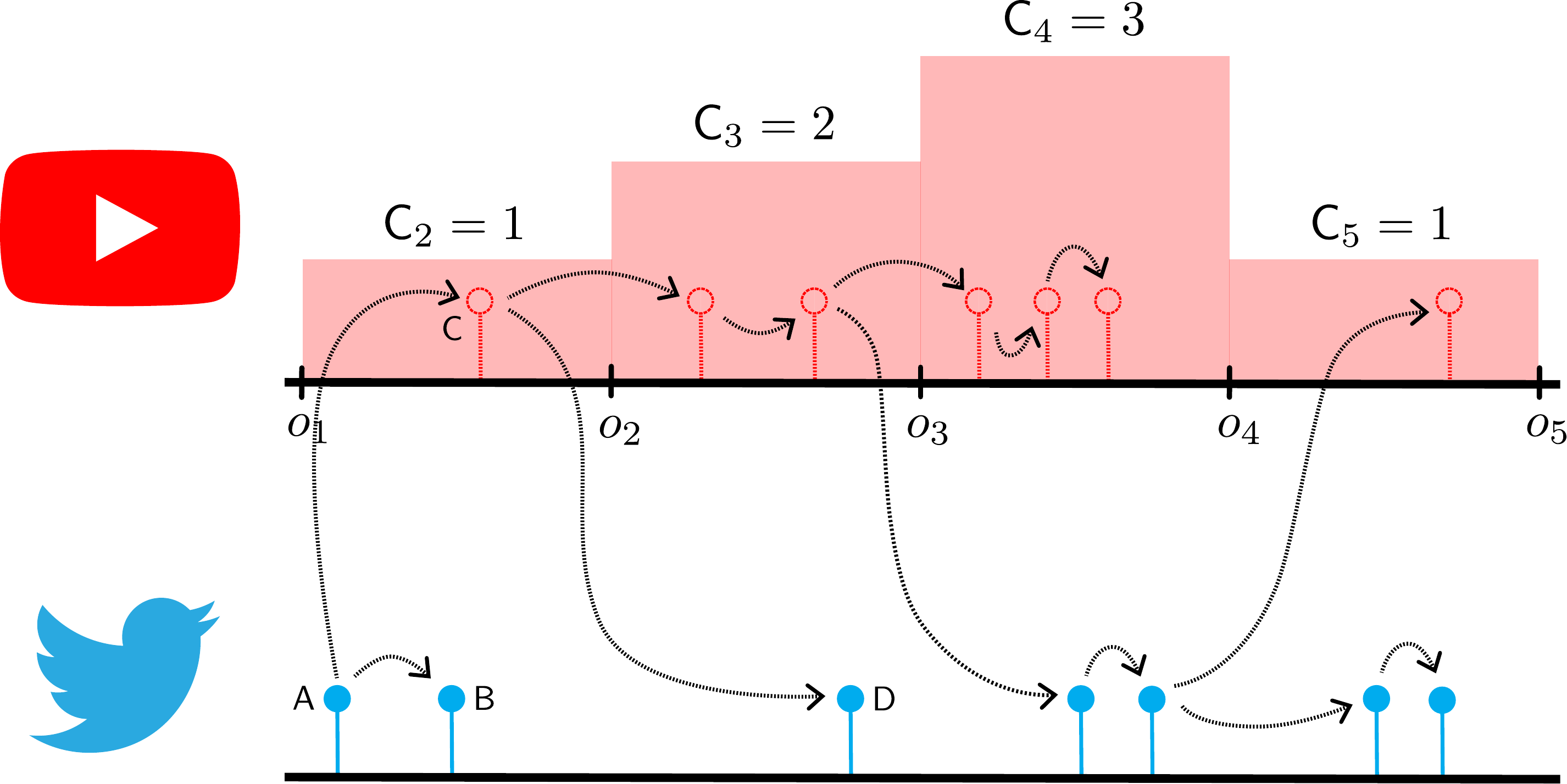}
    \caption{
        \textbf{Example of multi-platform interaction} between view events on YouTube  (\textcolor{red}{red lollipops}) and tweets on Twitter (\textcolor{blue}{blue lollipops}).
        The data is partially interval-censored, as YouTube does not expose individual views, but only the view counts $\HPObserVolume_i$'s over the predefined intervals $[o_i, o_{i+1})$ (shown as \textcolor{red}{red rectangles}). 
        The dashed lines show the latent branching structure between views and tweets.
        The red lollipops are also dashed and empty, indicating that YouTube views are not observed.
    }
    \label{fig:teaser}
\end{figure}

\IEEEPARstart{T}{he} Hawkes process, introduced by \cite{Hawkes1971}, is a temporal point process that exhibits the \emph{self-exciting} property, \ie, the occurrence of one event increases the likelihood of future events. 
The Hawkes process is widely applied in both the physical and social sciences.
For example, earthquakes are known to be temporally clustered: the mainshock is often the first in a sequence of subsequent aftershocks. 
In online social media, tweets by influential users typically induce cascades of retweets as the message diffuses over the social network \cite{Rizoiu2017T}.
The multivariate Hawkes process (MHP) \cite{Hawkes1971} extends the univariate process by allowing events to occur in multiple parallel timelines --- dubbed as \emph{dimensions}.
These dimensions interact via \emph{cross-excitation}, \ie, events in one dimension can spawn events in the other dimensions.
\cref{fig:teaser} schematically exemplifies the interaction between two social media platforms: YouTube and Twitter.
An initial tweet (denoted as A on the figure) spawns a retweet (B) via self-excitation and a view (C) via cross-excitation.
The cross-excitation goes both ways: the view C generates the tweet D.

Given the event timestamps, we can fit the parameters of the Hawkes process using maximum likelihood estimation (MLE). 
However, in many practical applications, the event times are not observed, and only counts over predefined time partitions are available.
We denote such data as \emph{interval-censored}.
For multivariate data, we denote the case when all dimensions are interval-censored as \emph{completely interval-censored}.
If only a subset of dimensions is interval-censored, we have \emph{partially interval-censored} data.

One reason for interval-censoring is data availability --- for epidemic data \cite{Browning2021}, we usually observe the aggregated daily counts of reported cases instead of detailed case information. 
Another reason is space limitations --- for network traffic data \cite{Shlomovich2020}, storing high-resolution event logs is impractical; they are stored as summaries over bins instead. 
A third reason is data privacy. 
%% MAR: we already talk about the red bars and the intervals in the caption of the figure. Not needed here!
This is the case for YouTube, as shown in the upper half of \cref{fig:teaser}, where the individual views are interval-censored, and we only observe aggregated daily counts.

This paper tackles three open questions about using the MHP with partially interval-censored data. 
The first question relates to fitting the process to both event time and interval-censored data.
When the data is presented as event times, the MHP can be fitted using MLE \cite{Daley2003}.
However, if the data is partially or completely interval-censored, MLE cannot fit the MHP process parameters because it lacks the independent increments property \cite{Rizoiu2022}.
Given interval-censored counts, one could approach fitting the Hawkes process naïvely by sampling event times uniformly over the intervals \cite{Unwin2021}.
However, this quickly hits scalability issues for high interval-censored counts. 
For instance, the Youtube videos in our real-world dataset often have millions of views per day.
For completely interval-censored univariate data, \cite{Rizoiu2022} proposed the Mean Behavior Poisson (MBP) --- an inhomogeneous Poisson process that approximates the mean behavior of the Hawkes process --- to estimate the parameters of a corresponding Hawkes process.
However, a model and fitting scheme remained elusive for the partially interval-censored data.
The question is, \textbf{can we devise a method to fit the MHP in the partially interval-censored setting?}
\textbf{What are the limits to MHP parameter recovery in the partially interval-censored setting?}

The second question relates to modeling and forecasting online popularity across social media platform boundaries.
Online popularity has been extensively studied within the realm of a single social media platform --- see Twitter~\cite{Zhao2015,Kobayashi2016,Mishra2016,Zadeh2022}, YouTube~\cite{Crane2008,Rizoiu2017}, Reddit~\cite{Krohn2019} --- and the self-exciting point processes are the tool of choice for modeling.
However, content is often shared across multiple interacting platforms --- such as YouTube and Twitter --- and we need to account for cross-excitation using multivariate processes.
However, YouTube only exposes view data as interval-censored, rendering it impossible to use the classical MHP.
The Hawkes Intensity process (HIP) \cite{Rizoiu2017} proposes a workaround and treats the tweet and share counts as external stimuli for views.
Its shortcoming is that it cannot model the cross-excitation from views to tweets and shares.
The question is, \textbf{can we improve performance in the YouTube popularity prediction task by modeling the views, tweets, and shares through fitting on partially interval-censored data?} 

The third question concerns analyzing interaction patterns across the online and offline environments, enabling us, for example, to determine whether online activity preempts or reacts to events that happen offline.
Previous work has demonstrated the complex link between news and infectious disease outbreaks, notably the 2009 A/H1N1 outbreak in the Shaanxi province in China \cite{yan2016}, the 2010 cholera outbreak in Haiti \cite{chunara2012}, and the early spread of COVID in 2020 in various provinces in China \cite{yan2020}. 
The association between media and case counts has typically been investigated by examining the cross-correlation of the news counts and case counts as paired time series and demonstrating that significant correlations exist when temporal lags are applied. 
\cite{yan2016,yan2020} show correlations between news and cases for both positive and negative lags, suggesting that news both had an impact and had been impacted by reported disease counts. 
\cite{chunara2012} show that news typically lags behind cases;
they also showcase how news counts can be used as a proxy for estimating crucial disease measures such as the basic reproduction number $R_0$.
This highlights that the connection between news and cases is particularly relevant given that news counts can be retrieved in near real-time; in contrast, official case counts reporting is often lagging.
% Other prior work \cite{Ortiz2020, Ciaffi2020, Strzelecki2020, Walker2020} have studied the relationship between Google Trends, a proxy for internet search activity, and COVID-19 incidence. 
In most previous work, uncovering time-series cross-correlation is the focus, without building explanatory models to produce nuanced views of the interactions through interpretable parameters.
The question is, \textbf{can we apply MHP on partially interval-censored data to uncover country-level differences in the interplay between recorded daily case counts of COVID-19 and the publication of COVID-19-related news articles?} 

We address these three questions by introducing the Partially Censored Multivariate Hawkes Process (\pmbp)\footnote[1]{Implementation available at \url{https://github.com/behavioral-ds/pmbp_implementation}.}, a novel multivariate temporal point process that operates on partially interval-censored data. 
We answer the first question in \cref{subsection:pmbp}, where we detail the \pmbp. 
The event intensity of \pmbp on the interval-censored dimensions is determined by the expected Hawkes intensity, considering the stochastic history of those dimensions conditioned on the event time dimensions. 
On the event time dimensions, the intensity of \pmbp corresponds to that of the respective Hawkes process. 
This construction allows us to fit the \pmbp to partially interval-censored data and estimate the parameters of the multivariate Hawkes process through parameter equivalence.

We address the second question in \cref{subsection:popularityprediction} by using \pmbp to predict the popularity of YouTube videos on both YouTube and Twitter.
We demonstrate that \pmbp consistently outperforms the related HIP method \cite{Rizoiu2017}, provides quantification of prediction uncertainty and extends predictions to all dimensions --- unlike HIP, which can only predict the views' dimension.

We address the third question in \cref{subsection:covid} by utilizing \pmbp to investigate the relationship between COVID-19 case incidence and news coverage. 
We fit a country-specific \pmbp for each of the 11 countries using a dataset consisting of reported COVID-19 cases (with interval-censored data) and the publication dates of COVID-19-related news articles during the early stage of the outbreak. 
We identify three distinct groupings by clustering countries using the fitted \pmbp parameters. 
In the first group (UK, Spain, Germany, and Brazil), we observe preemptive news coverage, where an increase in news leads to a rise in cases. 
The second group (China and France) exhibits reactionary news coverage, with news lagging behind the cases. 
No significant interaction between news and cases is found in the third group (US, Italy, Sweden, India, and the Philippines).

\subsection{Related Work}

A significant portion of recent literature on the Hawkes process, and on point processes in general, deals with  estimation from partially observed data. This problem is nontrivial as standard MLE techniques require the complete dataset. 

It was shown in \cite{Kirchner2016} that a sequence of (integer-valued autoregressive time series) INAR($\infty$)-based family of point processes converges to the Hawkes process. Under this convergence, they concluded that the INAR($\infty$) is the discrete-time version of the Hawkes process. In a follow-up, \cite{Kirchner2017} presented an alternative procedure to MLE, which fits the associated bin-count sequences to the INAR($p$) process. As the bin size goes to zero and the order $p$ of the process goes to $\infty$, the INAR sequence converges to the Hawkes process and the parameter estimates converge to the Hawkes parameters. However, though fitting is performed on count data, convergence only actually occurs for small bin size.

 A spectral approach to fitting the Hawkes process given interval-censored data for arbitrary bin size is presented in \cite{Cheysson2020}, solving the issue in \cite{Kirchner2017}. Their proposed method is based on minimizing the log-spectral likelihood of the bin-count sequence instead of the usual log-likelihood of the Hawkes process. They showed that optimization converges to the Hawkes parameters under certain assumptions on the kernel. 

The sample-based Monte Carlo Expectation Maximization (MC-EM) algorithm was introduced in~\pioc{\cite{Shlomovich2020} and \cite{Shlomovich2021} for the univariate and multivariate cases, respectively, which uses sampling to obtain proposals for the hidden event times. They showed that their approach recovers parameters more reliably than the INAR(p) estimates from \cite{Kirchner2017} in synthetic experiments. Another sample-based approach, the recursive identification with sample correction (RISC) algorithm, was introduced in \cite{Schneider2023}, where synthetic sample paths are iteratively generated and corrected to match the observed bin counts. Reliance on sampling makes these approaches more computationally expensive than the others.}

\pioc{Several modifications have been proposed to estimate the Hawkes process from daily count data in the context of modeling the spread of COVID \cite{Bertozzi2020}. A Hawkes process incorporating spatio-temporal covariates was estimated using an EM algorithm in \cite{Chiang2020}, while the least-squares approach was utilized in \cite{Schoenberg2023} to model state-level differences in transmission rates in the U.S.. A discrete-time Hawkes process for country-level COVID transmission was introduced in \cite{Browning2021} and fit using Bayesian inference.}

% \pioc{The notion of interval-censored data also appears in other contexts, most prominently in the analysis of failure time data. By \textit{interval-censored} data, what we mean here is that the failure time of interest of an observed process cannot be observed exactly and we only know that it lies within an interval or follow-up window, which is common in medical or public health studies involving clinical trials. This is different from what we consider in this work, where \textit{interval-censored} data means event counts over predefined time intervals instead of the events themselves, as exemplified for Youtube in \cref{fig:teaser}.}

\pioc{The notion of interval-censored data is also used in other fields, most prominently in the study of time-to-event (failure time) data \cite{Sun2006,Chen2012,Bogaerts2017}. 
In these works, `interval-censored' refers to situations where the precise time of an event of interest in an observational study is unknown; instead, we only know that it occurred within a certain window or follow-up period \cite{Du2021}. Meanwhile `partly interval-censored' refers to situations where some failure times are exactly observed while others are only known to lie within certain intervals \cite{Chen2023}.
This data type is prevalent in health and clinical research \cite{Ma2014}, where exact event times may not be directly observable due to the nature of study designs. 
Contrary to this, our work adopts the definition of `interval-censored' as outlined in \cite{Rizoiu2022,Kong2023}, where event times are inaccessible and we instead observe event counts over predefined time intervals, as exemplified by Youtube views in \cref{fig:teaser}.
Furthermore, in this work, we define partial censoring of a multivariate process as the censoring of specific dimensions, with the rest being observed as point processes.}
% !TEX root=../main.tex
%
\section{Model}

\pioc{For convenience, the list of notation that we use in this work is provided in \cref{tab:notation}.}

%!TEX root=../main.tex
%
\begin{table}[tbp]
    \centering
    \caption{ Important notation used in this paper.}
    \label{tab:notation}
    \setlength{\tabcolsep}{2.1pt}
    \begin{tabular}{cl}
        \toprule
        Symbol & Meaning \\
        \midrule
        $\boldsymbol{\alpha}$ & Hawkes branching matrix \\
        $\rho(\boldsymbol{\alpha})$ & spectral radius of $\boldsymbol{\alpha}$\\
        $f^{ij}(t)$ & exponential kernel from dimension $j$ to $i$ \\
        $\boldsymbol{\theta}$ & exponential kernel decay parameter matrix \\
        $\boldsymbol{\mu}(t)$ & deterministic background intensity \\
        $\boldsymbol{\varphi}(t)$ & Hawkes kernel, where $\varphi^{ij}(t) = \alpha^{ij} f^{ij}(t)$\\
        $\boldsymbol{\lambda}(t)$ & MHP conditional intensity\\
        ${\xi}(t)$ & MBP conditional intensity\\
        $D$ & overall set of dimensions for \pmbp($d,e$) \\
        $E$ & set of MBP dimensions for \pmbp($d,e$)  \\
        $E^c$ & set of Hawkes dimensions for \pmbp($d,e$)  \\
        $\HSf^j_{t^-}$ & event sequence history on dimension $j \in D$ \\
        $\HSf^A_{t^-}$ & union of event sequence histories on $A \subset D$ \\
        $\boldsymbol{\xi}_E(t)$ & conditional intensity for  \pmbp($d,e$)\\
        $\boldsymbol{\Xi}_E(t)$ & compensator for \pmbp($d,e$)\\
        $\boldsymbol{\Theta}$ & parameter set for \pmbp($d,e$)\\
        $T$ & terminal time \\
        $\LCal \left( \boldsymbol{\Theta}; T \right)$ & log-likelihood function for \pmbp($d,e$) \\
        $\LCal_{ \boldsymbol{\Theta}} \left( \boldsymbol{\Theta}; T \right)$ & gradient of log-likelihood function for \pmbp($d,e$) \\
        $\Delta^P$ & time axis partition length for numerical convolution\\
        $\gamma^h$ & convergence threshold for infinite sum truncation\\
        \bottomrule
    \end{tabular}
\end{table}
% !TEX root=../main.tex
%
\subsection{Background}
\label{section:background}

A temporal point process can be specified by its conditional intensity function. \pioc{In this work, we consider \textit{simple point processes}, where no two events can occur simultaneously.} Let $\textbf{\textsc{N}}(t)$ represent the number of events that occurred up until time $t$ and $\HSf^{j}_{t^-}$ be the set of all events that occur in dimension $j$ up until $t$, for $j \in \{1, \ldots, d\}$. \alexs{We further denote the union of all history dimensions as $\HSf_{t^-} \defeq \bigcup_{j=1}^d \HSf^{j}_{t^-}$.} The \pioc{$d$-dimensional} conditional intensity function $\boldsymbol{\lambda}\left(t | \HSf_{t^-}\right)$ is defined as
\begin{equation*}
    \lambda^j\left(t \alexs{\mid \HSf_{t^-}}\right) = \lim_{h \rightarrow 0^+} \frac{1}{h} \Pr \left\{ N^j(t+h) - N^j\alexs{(t)} = 1 \alexs{\mid \HSf_{t^-}}\right\},
\end{equation*}
which gives the instantaneous probability of a dimension $j$ event occurring in the increment $[t, t+dt)$, conditioned on all events that happen before $t$. \pioc{For brevity and whenever it is clear from context, we drop the explicit conditioning on the history  $\HSf_{t^-}$ and write $\boldsymbol{\lambda}(t) \defeq \boldsymbol{\lambda}\left(t | \HSf_{t^-}\right)$.}

{\bf{Hawkes Process:}} The univariate Hawkes process is a type of temporal point process that models a sequence of events on a single dimension exhibiting a \textit{self-exciting} behavior. Given $d$ types of events, the corresponding $d$-dimensional multivariate Hawkes process (MHP) is a point process where each dimension tracks the dynamics of each event type. In addition to being self-exciting, the MHP is \textit{cross-exciting} among event types, \ie, 
an event occurring in one type of event increases the probability of any type of event occurring in the near future.
The conditional intensity of the $d$-dimensional Hawkes process is given by
\begin{equation}
    \boldsymbol{\lambda}(t) \defeq \boldsymbol{\mu}(t) + \sum_{j=1}^d \sum_{t_k^j \in \HSf^j_{t^-}} \boldsymbol{\varphi}^j(t-t^j_k), \label{eqn:hawkes}
\end{equation}
where $\boldsymbol{\mu}(t)$ is the \pioc{(deterministic)} background intensity, \pioc{a $d$-dimensional non-negative vector for each $t$}  controlling the arrival of external events into the system. The matrix $\boldsymbol{\varphi}(t)$ is called the Hawkes kernel, a $d \times d$ matrix of functions that characterizes the self- and cross-excitation across the event types representing the $d$ dimensions. Let $\boldsymbol{\varphi}^j(t)$ represent the $j^{th}$ column of the Hawkes kernel. The diagonal entries $\varphi^{jj}(t)$ and off-diagonal entries $\varphi^{ij}(t)$, $i \neq j$, represent the self- and cross-exciting components of the Hawkes kernel, respectively. 
\pioc{Note that the Hawkes intensity $\boldsymbol{\lambda}(t)$ defined in \cref{eqn:hawkes} is a stochastic function dependent on the history of any particular realization $\HSf_{t^-}$. Given a fixed  $\HSf_{t^-}$, the intensity before or equal to $t$ is deterministically calculated using \cref{eqn:hawkes}. On the other hand, the intensity is a random variable for any time greater than $t$ or if the history $\HSf_{t^-}$ itself is not observable, such as in the case of interval censoring.}

The Hawkes kernel is often specified in a parametric form to facilitate simple interpretability. Let $D$ denote the index set $\{1, \ldots, d\}$. If we assume $\varphi^{ij}(t) = \alpha^{ij}f^{ij}(t) $, $ \alpha^{ij} \geq 0$, $ f^{ij}(t) \geq 0$, and $\int^{\infty}_{0} f^{ij}(t) \d t = 1 $ for $(i,j) \in D \times D$. We call $\alpha^{ij}$ the branching factor from $j$ to $i$ and the matrix $\boldsymbol{\alpha} = (\alpha^{ij}) \in (\Real^+)^{d \times d}$ the branching matrix. 
% The branching factor $\alpha^{ij}$ gives the expected number of direct offsprings in dimension $j$ spawned by a parent of dimension $i$.
\pioc{The branching factor $\alpha^{ij}$ gives the expected number of offspring events in dimension $i$ that are triggered by an event in dimension $j$.}
\pioc{The function $f^{ij}(t)$ is typically selected to be monotonically decreasing to model the empirically observed decay in the attention that online content receives over time \cite{Crane2008}. 
This is explained by viewing human attention as a limited resource that online content competes for, resulting in content being forgotten over time.} 
In this work we consider the widely used exponential kernel \cite{Shen2014,Mishra2016,Rizoiu2022}, which takes the form $\varphi^{ij}(t) = \alpha^{ij} \theta^{ij} \exp(-\theta^{ij} t)$, where $\theta^{ij}$ controls the rate of influence decay from $j$ to $i$. 
We choose the exponential kernel due to its simplicity, enabling efficient computation of the intensity due to its Markovian property \cite{Ogata1981}, efficient sampling \cite{dassios2013exact}, and strong theoretical properties, as evidenced by its link to the stochastic SIR model \cite{rizoiu2018sir}. 
An alternative choice of kernel is the power law, which, while yielding better model likelihood, requires significantly more time to fit and presents a more challenging estimation due to the need to fit two parameters. 
More prerequisite details on the MHP are provided in Sec. 1.1 of the SI \cite{appendix}.

{\bf{Mean Behavior Poisson Process~\cite{Rizoiu2022}:}} Consider a univariate Hawkes process with conditional intensity $\lambda(t)$. The Mean Behavior Poisson (MBP) process introduced by \cite{Rizoiu2022} is the inhomogeneous Poisson process with conditional intensity
\begin{equation}
    \xi(t) \defeq \E{\HSf_{t^-}}{\lambda(t)}. \label{eq:mbp_def}
\end{equation}
\pioc{In contrast to the stochastic Hawkes intensity $\lambda(t | \HSf_{t^-})$, the MBP intensity $\xi(t)$ is a deterministic function obtained by taking the expectation of the Hawkes intensity over all possible realizations $\{\HSf_{t^-}\}$.} \alexs{It was shown in \cite{Rizoiu2022} that $\xi(t)$ follows the self-consistent equation}
\begin{equation}
    \xi(t) = \mu(t) +  (\varphi * \xi)(t) \label{eq:mbp_intensity_convolution},
\end{equation}
where $*$ denotes convolution. Furthermore, the mapping $\mu(t) \Mapsto \xi(t)$ in \cref{{eq:mbp_intensity_convolution}} defines a linear time-invariant (LTI) system \cite{Phillips2003}, meaning that it obeys linearity ($\mu_1(t) \Mapsto \xi_1(t)$ and $\mu_2(t) \Mapsto \xi_2(t)$ imply that $a \mu_1(t) + b \mu_2(t) \Mapsto a \xi_1(t) + b \xi_2(t)$ for $a, b \in \mathbb{R}$) and time invariance ($\mu(t) \Mapsto \xi(t)$ implies that $\mu(t-t_0) \Mapsto \xi(t-t_0)$ for $t_0 > 0$.) As an LTI system, the response $\xi(t)$ to the input $\mu(t)$ can be obtained by solving for the response of the system to the Dirac impulse $\delta(t)$, derived in \cite{Rizoiu2022} to be
\begin{equation}
    \label{eqn:mbp_intensity}
    \xi(t) = \left(\delta(t) + \sum_{n=1}^\infty \varphi^{\otimes n}(t)\right) * \mu(t),
\end{equation}
where $\otimes n$ corresponds to $n$-time self-convolution.

Since the MBP process is a Poisson process, its increments are independent, which allows the likelihood function to be expressed as a sum of the likelihood of disjoint Poisson distributions. This enables the MBP process to be fitted in interval-censored settings via MLE. More prerequisite details are provided in Sec. 1.2 of the SI \cite{appendix}.

{\bf{Hawkes Intensity Process~\cite{Rizoiu2017}:}} The Hawkes intensity process (HIP), introduced in \cite{Rizoiu2017}, is a temporal point process that can be fit to interval-censored data. It was used primarily for YouTube popularity prediction, where YouTube video views are daily-censored, and external shares and tweets that mention the video act as the exogenous intensity $\mu(t)$.

Given a partition $\mathcal{P}[0,T) = \bigcup_{k=1}^{m} [o_{k-1}, o_{k})$, where $o_0 = 0$ and $o_m = T$, and the associated view counts $\{\HPObserVolume_k\}_{k=1}^m$, the HIP model \alexs{$\hat{\xi}[\cdot; \boldsymbol{\Theta}]$} is fitted by finding the parameter set $\boldsymbol{\Theta}$ that minimizes the sum of squares error $\sum_{k=1}^m \left( \HPObserVolume_k - \hat{\xi}[o_k; \boldsymbol{\Theta}]\right)^2$
\pioc{of the following recursive formula for $\hat{\xi}[o_k; \boldsymbol{\Theta}]$,}  %$\hat{\xi}[\cdot]$ is given by
\begin{equation*}
    \hat{\xi}[o_k; \boldsymbol{\Theta}] = \mu[o_k] + \sum_{s=0}^{k-1} \varphi(o_k - o_s; \boldsymbol{\Theta}) \cdot \hat{\xi}[o_s; \boldsymbol{\Theta}].
\end{equation*}
The use of brackets emphasizes that the quantities are discretized over a partition of time.
It was shown in \alexs{\cite[Theorem 10]{Rizoiu2022}} that HIP is a discretized approximation of the MBP process\alexs{, where an implicit assumption that the observation intervals being unit length is reflected in the sum of squares error, \ie, \( \hat{\xi}[o_k; \boldsymbol{\Theta}] \cdot (o_k - o_{k-1}) \) is approximated as \( \hat{\xi}[o_k; \boldsymbol{\Theta}] \).}
% !TEX root=../main.tex
%
\subsection{Partially Censored Multivariate Hawkes Process}
\label{subsection:pmbp}

We define the Partially Censored Multivariate Hawkes Process \( \pmbp(d,e) \) with intensity $\boldsymbol{\xi}_E(t)$ as follows. \alexs{A key idea of the \( \pmbp(d,e) \) is to fix the history of different dimensions. As such we denote the history union over a subset of dimensions \( A \subset \{ 1 \ldots d \} \) as \( \HSf^{A}_{t^-} \defeq \bigcup_{j\in A} \HSf^{j}_{t^-} \).}

\begin{definition} 
\label{def:pmbp_definition}
Consider a $d$-dimensional Hawkes process with conditional intensity $\boldsymbol{\lambda}(t)$ as defined in \cref{eqn:hawkes}. Given a nonnegative integer $e \leq d$ and the index sets $D  \defeq \{1, \ldots, d\}$, $E \defeq \{1, \ldots, e\}$ and $E^c \defeq \{ e+1, \ldots, d\}$, the Partially Censored Multivariate Hawkes Process \( \pmbp(d,e) \) is the temporal point process whose conditional intensity $\boldsymbol{\xi}_E(t)$ is the expectation of $\boldsymbol{\lambda}(t)$ conditioned on the set of event histories 
%$\bigcup_{j \in E^c}\HSf^{j}_{t^-}$
\alexs{$\HSf^{E^c}_{t^-}$}
in the $E^c$ dimensions and averaged over the set of event histories 
%$\bigcup_{j \in E}\HSf^{j}_{t^-}$
\alexs{$\HSf^{E}_{t^-}$}
in the $E$ dimensions. That is,
    \begin{equation}
        \label{eqn:pmbp_definition}
        \boldsymbol{\xi}_E(t) \defeq \boldsymbol{\xi}\left(t \alexs{\mid \HSf^{E^c}_{t^-}}\right) = \E{\alexs{\HSf^{E}_{t^-}}}{\boldsymbol{\lambda}(t) \alexs{\mid \HSf^{E^c}_{t^-}}}.
    \end{equation}        
\end{definition}

% The \( \pmbp(d,e) \) can be viewed as a collection of $d$ processes where the dynamics in the $E$ dimensions follow an $e$-dimensional inhomogeneous Poisson process and the $E^c$ dimensions are governed by a ($d-e$)-dimensional Hawkes process.
\pioc{The $\pmbp(d,e)$ intensity is a stochastic function due to its dependence on the current realization of $\HSf^{E^c}_{t^-}$; on the $E$ dimensions we take the expectation over all possible realizations of $\HSf^{E}_{t^-}$, similar to the MBP intensity in \cref{eq:mbp_def}.}

In practice, $E$ would be chosen to be the set of dimensions where event times are inaccessible and only interval-censored event counts can be obtained, while $E^c$ would be the dimensions with event time information. In \cref{fig:teaser} for instance $E$ would be YouTube views and $E^c$ the set of tweets.

\pioc{
{\bf{Is the \pmbp($d,e$) Poisson?}} Due to its dependence on the history of the $E^c$ dimensions, the \pmbp($d,e$) is not a Poisson process. From \cref{eqn:pmbp_definition}, the \( \pmbp(d,e) \) can be interpreted as a collection of $d$ processes, where $\xi^j_E(t)$ follows a Hawkes process for $j \in E^c$ and follows an inhomogeneous Poisson process (conditional on the event history of the $E^c$ dimensions) for $j \in E$. In fact, the \( \pmbp(d,e) \) generalizes both the MHP (by setting $e = 0$) and the MBP process (by setting $e = d$). 
}

{\bf{Convolutional Formula:}} Consider the kernel $\boldsymbol{\varphi}(t)= \left[
    \begin{array}{cccccc}
      \boldsymbol{\varphi}^1(t) & \ldots & \boldsymbol{\varphi}^e(t) & \boldsymbol{\varphi}^{e+1}(t) & \ldots & \boldsymbol{\varphi}^d(t)       \end{array}
  \right]$, setting $\boldsymbol{\varphi}^j(t)$ to be the $j^{th}$ column of $\boldsymbol{\varphi}(t)$.

Similarly, let $\boldsymbol{\varphi}_E(t) = \left[
    \begin{array}{cccccc}
      \boldsymbol{\varphi}^1(t) & \ldots & \boldsymbol{\varphi}^e(t) & 0 & \ldots & 0       \end{array}
  \right]$ and $\boldsymbol{\varphi}_{E^c}(t) = \left[
    \begin{array}{cccccc}
      0 & \ldots & 0 & \boldsymbol{\varphi}^{e+1}(t) & \ldots & \boldsymbol{\varphi}^d(t)\\
    \end{array}
  \right]$.

Similar to MBP, $\boldsymbol{\xi}_{E}(t)$ can be expressed as the response of an LTI system, which allows us to express $\boldsymbol{\xi}_{E}(t)$ as a convolution with the Dirac impulse $\boldsymbol{\delta}(t)$. 

\begin{theorem}
    \label{thm:impulse_response_formula}
    Given the Hawkes process with intensity \cref{eqn:hawkes} and kernel parameters satisfying $\lim_{n \rightarrow \infty} \boldsymbol{\varphi}_E^{\otimes n}(t) = 0$, the conditional intensity of the \( \pmbp(d,e) \) is
    \begin{equation}
        \label{eqn:impulse_response_formula}
        \boldsymbol{\xi}_{E}(t) = \left[ \boldsymbol{\delta}(t) + \sum^{\infty}_{n = 1} \boldsymbol{\varphi}_{E}^{\otimes n}(t) \right] * \left[ \boldsymbol{\mu}(t) + \sum_{j \in E^c} \sum_{t^j_k < t} \boldsymbol{\varphi}^j_{E^c} (t - t^j_k) \right],
    \end{equation}
\end{theorem}

In general, $\boldsymbol{\xi}_{E}(t)$ does not admit a closed form solution because of the complexity of the infinite convolution sum of $\boldsymbol{\varphi}_{E}(t)$ (an interpretation of which is provided in Sec. 2 of the SI \cite{appendix}). However, in the special case of $\pmbp(2,1)$ with the exponential kernel, a closed-form solution for $\boldsymbol{\xi}_{E}(t)$ exists, as proven in Sec. 3 of the SI. 

{\bf{Regularity Conditions:}} 
Imposing regularity conditions on the model parameters ensure process \textit{subcriticality}, \ie the expected number of direct and indirect offspring spawned by a single parent is finite. For instance, an MHP is subcritical if the spectral radius \pioc{$\rho$} (\ie magnitude of the largest eigenvalue) of the branching matrix is less than one, \ie $\rho(\boldsymbol{\alpha}) < 1$ \cite{Ogata1978}. Here we introduce the regularity conditions applicable for the  \( \pmbp(d,e) \).

Consider the following submatrices of $\boldsymbol{\alpha}$:
\begin{align*}
    \boldsymbol{\alpha}^{EE} &= (\alpha^{ij})_{(i,j) \in E \times E},& \boldsymbol{\alpha}^{E E^c} &= (\alpha^{ij})_{(i,j) \in E \times E^c},\\
    \boldsymbol{\alpha}^{E^c E} &= (\alpha^{ij})_{(i,j) \in E^c \times E},& \boldsymbol{\alpha}^{E^c E^c} &= (\alpha^{ij})_{(i,j) \in E^c \times E^c}.
\end{align*}

The following are three conditions which ensure subcriticality of \( \pmbp(d,e) \).

\begin{theorem}
    \label{thm:regularity_pmbp}
    The \( \pmbp(d,e) \) with branching matrix $\boldsymbol{\alpha}$ is subcritical if the following conditions hold.
    \begin{equation*}
        \rho(\boldsymbol{\alpha}^{EE}) < 1 \text{, } \rho(\boldsymbol{\alpha}^{E^c E^c}) < 1 \text{, } \rho(\boldsymbol{\alpha}^{E^c E} (\mathbf{I} - \boldsymbol{\alpha}^{EE})^{-1} \boldsymbol{\alpha}^{E E^c})  < 1.
    \end{equation*}
\end{theorem}

We note that the regularity conditions for \( \pmbp(d,e) \) in \cref{thm:regularity_pmbp} cover the MHP and the MBP as special cases.

Proofs of \cref{thm:impulse_response_formula} and  \cref{thm:regularity_pmbp} and a discussion on the nonlinear extension of the \pmbp are provided in Sec. 4 of the SI \cite{appendix}.

\subsection{Inference} 
\label{subsec:inference}

We now consider the problem of estimating the \(\pmbp\) parameter set $\boldsymbol{\Theta}$ given a \pioc{partially interval-censored} dataset consisting of interval-censored data on a subset of dimensions and exact event sequences on the other dimensions. \pioc{Assuming a constant exogenous term $\boldsymbol{\mu}(t) = \boldsymbol{\nu}$, the \(\pmbp\)($d,e$) parameter set to be estimated is given by $\boldsymbol{\Theta} = \{\boldsymbol{\nu}, \boldsymbol{\alpha}, \boldsymbol{\theta}\}$ with size $|\boldsymbol{\Theta}| = d + 2 \cdot d^2$.}

Consider a $d-$dimensional dataset over the time interval $[0, T)$ such that observations in the first $q$ dimensions are interval-censored, and in the last $d-q$ dimensions, we observe event times. 
Formally,
let $Q \defeq \{1, \ldots, q\}$ and $Q^c \defeq \{q+1, \ldots, d\}$. 
For $ j \in Q$, we associate a set of observation points $o_0^j < o_1^j < \ldots < o_{n^j}^j$ such that for $o_k^j$ where $k\geq 1$, we observe the volume $\HPObserVolume^j_k$ of dimension $j$ events that occurred during the interval $[o_{k-1}^j, o_k^j)$. 
Meanwhile, for $j \in Q^c$, we observe event sequences $\HSf_{T^-}^j = \{ t^j_1 < t^j_2 < \ldots < t^j_{n^j}\}$.
\pioc{In practice, the observation partition $\bigcup_{j \in Q} \bigcup_{k=1}^{n^j} [o_{k-1}^j, o_k^j)$ is not a model hyperparameter but is determined by real-world dataset availability constraints. For instance, in \cref{fig:teaser} we consider a daily partitioning for Youtube views since our dataset consists of aggregated daily view counts.
}

We use MLE to fit the parameters of a \(\pmbp(d,e)\) process to the above-defined data using 
the log-likelihood function derived below. The proof is available in Sec. 5 of the SI \cite{appendix}.

\begin{theorem} \label{th:PMBP-likelihood}
    Given event times 
    %\bigcup_{j \in Q^c} \HSf_{T^-}^j$
    \alexs{$\HSf_{T^-}^{Q^c}$}, event volumes $\bigcup_{j \in Q} \{\HPObserVolume^j_k\}_{k = 1}^{n^j}$, and a \( \pmbp(d,e) \) model such that $E \supseteq Q$, the negative log-likelihood of parameter set $\boldsymbol{\Theta}$ can be written as
    \begin{equation}
    \label{eqn:pmbp_nll}
        \LCal \left( \boldsymbol{\Theta}; T \right) = \sum_{j \in Q} \npll^j \left( \boldsymbol{\Theta}; T \right) + \sum_{j \in Q^c} \ppll^j \left( \boldsymbol{\Theta}; T \right),
    \end{equation}
    where
    \begin{align}
        \label{eqn:pmbp_nll_icll}
        \npll^j \left( \boldsymbol{\Theta}; T \right) &= \sum_{i=1}^{n^j} \left[ \MBPcomp_E^j(\obsTime_{i-1}^j, \obsTime_{i}^j; \boldsymbol{\Theta}) - \HPObserVolume_i^j \log \MBPcomp_E^j(\obsTime_{i-1}^j, \obsTime_{i}^j; \boldsymbol{\Theta}) \right], \\
        \label{eqn:pmbp_nll_ppll}
        \ppll^j \left( \boldsymbol{\Theta}; T \right) &= - \sum_{t_k^j \in \HSf_{T^-}^j} \log \xi_E^j(t^j_k; \boldsymbol{\Theta} ) + \Xi_E^j(T; \boldsymbol{\Theta} ),
    \end{align}
and $\boldsymbol{\MBPcomp}_E(t)$ represents the compensator, \ie, the intensity $\boldsymbol{\xi}_E(t)$ integrated over $0$ to $t$.
\end{theorem}

{\bf{Choice of Likelihood:}}
The choice of likelihood on a given dimension $j$ is solely dependent on the type of data on the said dimension.
If \(j \in Q\) (dimension $j$ is interval-censored), one should use \(\npll^j \left( \boldsymbol{\Theta}; T \right)\); 
if \(j \in Q^c\) (event-times) then \(\ppll^j \left( \boldsymbol{\Theta}; T \right)\) should be used. 

An event-time dimension ($j \in Q^c$) can be modeled using either the Hawkes dynamics or the MBP dynamics.
However, an interval-censored dimension ($j \in Q$) can only be modeled using MBP dynamics, as an interval-censored log-likelihood for the Hawkes dynamics does not exist.
It follows that \(E \supseteq Q\). 
In real-world applications, one would choose \(E = Q\) because any other choice \(E \supset Q\) leads to information loss due to the mismatch between the data generation model (\ie, Hawkes) and the fitting model (MBP). 
We study the impact of model mismatch loss in \cref{subsection:parameterrecovery}.

{\bf{Runtime Complexity:}}
Denote $\text{n}^{E^c}$ and $\text{n}^{Q^c}$ as the total number of observed event times in the $E^c$ and $Q^c$ dimensions, respectively; and $\text{n}^{E}$ and $\text{n}^{Q}$ as the total number of observation intervals in the $E$ and $Q$ dimensions, respectively. That is, 
$\text{n}^{E^c} = \sum_{j \in E^c} \vert\HSf_{T^-}^j\vert$, $\text{n}^{Q^c} = \sum_{j \in Q^c} \vert\HSf_{T^-}^j\vert$,
$\text{n}^{E} = \sum_{j \in E} {n^j}$, and
$\text{n}^{Q} = \sum_{j \in Q} {n^j}$.
Let \( C \) denote a constant independent of the dimension of the \pmbp and the data.
Evaluating $\LCal(\boldsymbol{\Theta};T)$ has a runtime complexity of $\mathcal{O}\left((C + \text{n}^{E^c}) \cdot (\text{n}^{Q} + \text{n}^{Q^c})\right)$ (see Sec. 5 of the SI \cite{appendix} for more details).
In the case $E=Q=\emptyset$, the runtime complexity reduces to $\mathcal{O}(({\text{n}^{E^c}})^2)$, consistent with the MHP. If $E=Q=D$, runtime complexity reduces to $\mathcal{O}({\text{n}^{E}})$, consistent with the MBP (i.e. Poisson) process.

{\bf{Numerical Considerations:}}
\pioc{Due to the complexity of $\sum^{\infty}_{n = 1} \boldsymbol{\varphi}_{E}^{\otimes n}(t)$ and its convolutions, a general closed-form expression for $\boldsymbol{\xi}_E(t)$ is not available, requiring us to leverage approximation techniques, \ie numerical convolution and infinite series truncation, to compute $\sum^{\infty}_{n = 1} \boldsymbol{\varphi}_{E}^{\otimes n}(t)$ and $\boldsymbol{\xi}_E(t)$. 
The approximation error is controlled by two hyperparameters: (1) $\Delta^P$, the partition length of our time axis for the numerical convolution, and (2) $\gamma^h$, the max-norm convergence threshold to determine $k^* \in \mathbb{N}$ to truncate the infinite sum, i.e. $\sum^{k^*}_{n = 1} \boldsymbol{\varphi}_{E}^{\otimes n}(t) \approx \sum^{\infty}_{n = 1} \boldsymbol{\varphi}_{E}^{\otimes n}(t)$.    
The smaller $\Delta^P$ and $\gamma^h$ are set, the tighter the approximation, albeit with a longer computation time.
Full details and heuristics on hyperparameter choice are discussed in Sec. 7 of the SI \cite{appendix}. 
We propose an alternative sampling-based technique to calculate $\boldsymbol{\xi}_E(t)$ that bypasses calculation of $\sum^{\infty}_{n = 1} \boldsymbol{\varphi}_{E}^{\otimes n}(t)$ in Sec. 8 of the SI \cite{appendix}. 
Finally, we demonstrate the convergence of the numerical and the sampling-based approximation techniques in Sec. 9 of the SI \cite{appendix} by showing close agreement between the approximated $\boldsymbol{\xi}_E(t)$ and the closed-form $\boldsymbol{\xi}_E(t)$ of the exponential \pmbp(2,1) derived in Sec. 3 of the SI \cite{appendix}.
}

Gradient-based optimization tools
-- including \textsc{IPOpt}~\cite{Wachter2006} that we use in our experiments in \cref{subsection:popularityprediction} and \cref{{subsection:covid}} --- usually require the gradient. 
To approximate $\LCal(\boldsymbol{\Theta};T)$ and its gradient $\LCal_{\boldsymbol{\Theta}}(\boldsymbol{\Theta};T)$, we propose a numerical scheme 
%with time complexity scaling as $\mathcal{O}([\text{n}^{E} + \text{n}^{E^c} + \lceil \frac{T}{\Delta^P} \rceil]\cdot \text{n}^{E^c} \cdot d)$ for $E^c \neq 0$. 
in Sec. 10 and 11 of the SI \cite{appendix}.
\pioc{We show in the SI that the runtime complexity of the numerical scheme is mostly determined by how many dimensions we model as Hawkes and as MBP. Without any Hawkes dimensions ($E^c = \emptyset$), the scheme scales linearly (similar to the MBP) with the number of observation intervals $n^E$, \ie $\mathcal{O} \left(\text{n}^{E} \cdot \lceil \frac{T}{\Delta^P} \rceil \cdot d \cdot e \right)$. On the other hand, if $E^c \neq \emptyset$, the scheme scales quadratically (similar to the MHP) with the number of observed event times $n^{E^c}$, \ie $\mathcal{O}([\text{n}^{E} + \text{n}^{E^c} + \lceil \frac{T}{\Delta^P} \rceil]\cdot \text{n}^{E^c} \cdot d)$. In both cases, the number of partition intervals $\lceil \frac{T}{\Delta^P} \rceil$ only appears linearly. For partially interval-censored datasets with high frequency data, the number of observed event times $n^{E^c}$ is the most important determinant of runtime complexity given that it appears quadratically, while the number of observation intervals $n^E$ and the number of partition intervals $\lceil \frac{T}{\Delta^P} \rceil$ only appear linearly.
}

Lastly, for the purpose of sampling from the \pmbp($d,e$), we propose a modification of the thinning algorithm \cite{Ogata1981} detailed in Sec. 12 of the SI.

\subsection{Heuristics for Partially Interval-Censored Data}
\label{subsection:discussion}

The \pmbp is designed for cases where (1) the dataset is multivariate and partially interval-censored, and (2) we hypothesize events are self-exciting within and cross-exciting across dimensions. To handle partially interval-censored data, our  strategy is to adapt the model (\ie the \pmbp) to the data. However, we can take the reverse approach and apply heuristics to our dataset  to be able to leverage pre-existing models.

\begin{enumerate}
    \item To use \textit{count-based time series models}, we transform our partially interval-censored dataset into a fully interval-censored dataset by censoring event times for each $E^c$ dimension.
    \item To use \textit{point process models (\ie the MHP)}, we transform our partially interval-censored dataset into a fully time-stamped dataset by sampling event times to match the interval-censored counts, for each dimension in $E$).
\end{enumerate}

There are three arguments against the first heuristic. First, artificially censoring the dataset leads to loss of timing information by hiding self- and cross-exciting interactions between events, particularly if the time scale of the interactions is less than the censor window length. Second, commonly used time series models (such as the Poisson autoregressive model \cite{Fokianos2009} or the discrete-time Hawkes process \cite{Browning2021}) assume evenly spaced data \cite{Eckner2012}. If the censor intervals within or across dimensions do not line up, we would need to perform further data alteration, such as interpolation \cite{Rehfeld2011}, to attain evenly spaced data. The \pmbp does not require evenly spaced intervals. Third, using time series models on the artificially obtained interval-censored dataset requires additional model choices. For instance, we would have to set the censor window length for each dimension in $E$ when transforming to a fully interval-censored dataset, and for autoregressive models decide up to what lag $p$ to include. The \pmbp requires no adaptation as it was designed for partially interval-censored data.

The main deterrent against the second heuristic, artificial event sampling, is the significant addition to computation time, since evaluating the Hawkes likelihood is $\mathcal{O}((\text{n}^{E^c})^2)$. This is particularly infeasible in applications involving high event volumes, such as Youtube views on a viral video, which typically have view counts of the order $10^6$ or more. If we use the \pmbp, these dimensions with high event volumes can be modeled as event counts and placed in $E$ instead of $E^c$, significantly reducing computation time. Second, artificially sampling points --- when only aggregated counts have been given --- has the potential to produce spurious event interactions across dimensions, particularly for wide censor intervals.
% !TEX root=../main.tex
%
\section{Synthetic Parameter Recovery}
\label{subsection:parameterrecovery}

In this section, we test on synthetic data the MHP parameter recovery by \( \pmbp(d,e) \).
We use the setting of partial interval-censoring with a constant exogenous term $\boldsymbol{\mu}(t) = \boldsymbol{\nu}$. 
We sample realizations from a $d$-dimensional MHP, interval-censor $e$ dimensions using increasingly wide observation window lengths, and fit the \( \pmbp(d,e) \) model on the obtained partially interval-censored data.
We inspect the recovery of parameters when varying $d$ and $e$.
We perform convergence analysis on the \( \pmbp(d,e) \) parameter estimates for various hyperparameter configurations in Sec. 14.2 of the SI \cite{appendix},

Throughout this section, we refer to $\{\boldsymbol{\alpha}, \boldsymbol{\theta}, \boldsymbol{\nu}\}$ and $\{\hat{\boldsymbol{\alpha}}, \hat{\boldsymbol{\theta}}, \hat{\boldsymbol{\nu}}\}$ as the true (MHP) and estimated parameter sets, respectively. We first discuss the two types of information loss, then we introduce the synthetic datasets and the likelihood functions.
Lastly, we present the recovery results for the individual parameters $\{\boldsymbol{\alpha}, \boldsymbol{\theta}, \boldsymbol{\nu}\}$.

{\bf{Sources of Information Loss in Fitting:}}
We identify two sources of information loss when fitting in the partially interval-censored setup:
(1) the mismatch between the data-generation model (\ie, the $d$-dimensional MHP) and the fitting model (the \( \pmbp(d,e) \)); and 
(2) the interval-censoring of the timestamped MHP data. 
\pioc{Since the intensity $\boldsymbol{\xi}_E(t)$ of the \pmbp($d,e$) has to be estimated numerically (see \cref{subsec:inference}), numerical approximation error also contributes to type (1) information loss. The numerical approximation error is minimal for sufficiently small $\Delta^P$ and $\gamma^h$ (see Sec. 9 of the SI \cite{appendix}) and vanishes if $\Delta^P, \gamma^h \rightarrow 0$.}

When we estimate MHP parameters using $\pmbp$ fit on partially interval-censored data, information losses of both types (1) and (2) occur.
We disentangle between the two types of error by also fitting \( \pmbp(d,e) \) on the timestamp dataset (\ie, the actual realizations sampled from the MHP, see below).
Any information loss in this setup is only due to the model mismatch, the information loss of type (1).
Note that the likelihood function used for fitting $\pmbp$ depends on the employed version of the dataset (see later in this section).

We can quantify the individual effects of model mismatch and interval-censoring by comparing the parameter estimates on the two dataset versions.

{\bf{Synthetic Dataset:}} 
Given $(d,e)$, we construct two synthetic datasets: 
the \emph{timestamp dataset} and the \emph{partially interval-censored dataset}.
\pioc{The timestamp dataset consists of samples from a $d$-dimensional MHP.
The partially interval-censored dataset is obtained from the timestamp dataset by interval-censoring the $E$ dimensions while leaving the $E^c$ dimensions unchanged. 
In this experiment, we focus on the case $d=2$ and $e=1$.}

\begin{figure*}[t!]
    \includegraphics[width=\textwidth]{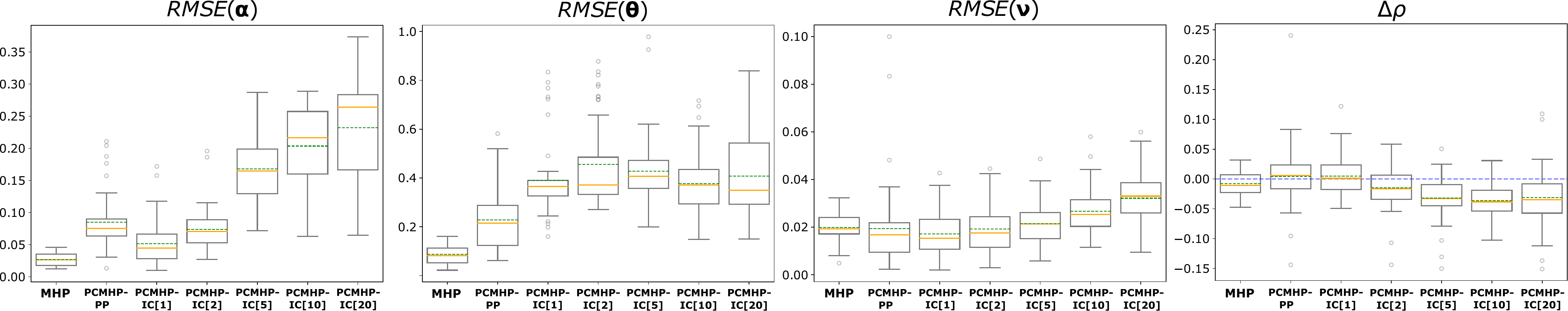}
    \caption{
        Comparison of performance metrics in the parameter recovery experiment across model fits: MHP (\ie the data-generating process), \pmbp-PP and \pmbp-IC for varying interval sizes (1, 2, 5, 10 and 20). (left to right) RMSE for each parameter type $\{\boldsymbol{\alpha}, \boldsymbol{\theta}, \boldsymbol{\nu}\}$ and spectral radius estimation error $\Delta \rho$.  
        Samples are drawn from a 2-dimensional MHP with spectral radius $\rho(\boldsymbol{\alpha}) = 0.75$.
        \pioc{Each boxplot represents the distribution of errors calculated from 50 estimates of $\boldsymbol{\Theta}$, each obtained from a different group in the synthetic dataset.}
        The mean and median estimates are indicated by the dashed green lines and solid orange lines, respectively.
    }
    \label{fig:err}
\end{figure*}

We consider a 2-dimensional MHP with $\rho(\boldsymbol{\alpha}) = 0.75$ and parameters $\alpha^{11} = 0.32$, $\alpha^{12} = 0.5$, $\alpha^{21} = 0.3$, $\alpha^{22} = 0.4$, $\theta^{11} = 0.5$, $\theta^{12} = 1.0$, $\theta^{21} = 0.5$, $\theta^{22} = 1.25$ and $\nu^1 = \nu^2 = 0.1$
We set $\rho(\boldsymbol{\alpha}) \in \{0.5, 0.75, 0.9\}$. 
We also test another parameter combination with $\rho(\boldsymbol{\alpha}) = 0.5$ (\ie subcritical) and $\rho(\boldsymbol{\alpha}) = 0.9$ (\ie approaching the critical regime) in Sec. 14.1 of the SI \cite{appendix}. 

For a given MHP parameter set, we sample $2500$ event sequences $\HSf^{1}_{100} \cup \HSf^{2}_{100}$ over the time interval $[0,100)$ using the MHP thinning algorithm \cite{Ogata1981}. 
Following a procedure similar to prior literature~\cite{Rizoiu2022},
we partition the 2500 event sequences into 50 groups of 50 sequences.
%, and each group of $N_{sequences}=50$ events is used for joint fitting, yielding a single parameter set estimate.
%In total, we obtain 50 sets of parameter estimates from the sample.

We construct the partially interval-censored dataset
by interval-censoring $\HSf^{1}_{100}$, the first dimension of each realization in the timestamp dataset.
Given a partition of $[0,100)$, we count the number of events on dimension $1$ that fall on each subinterval.
We experiment with five observation window lengths to quantify the information loss of type (2) -- intuitively, longer intervals lead to more significant information loss.
We consider interval lengths of $1$, $2$, $5$, $10$ and $20$. For instance, with interval length of $2$ we tally event counts in the partition $\{[0,2), [2,4), \ldots, [98,100)\}$. 

{\bf{\( \pmbp \) Log-Likelihood Functions:}}
We fit the parameters of the \( \pmbp(d,e) \) model using two different versions of the likelihood function dependent on which dataset we use:

\begin{itemize}
    \item \emph{timestamp dataset:} we use the point-process log-likelihood on all dimensions, defined in \cref{eqn:pmbp_nll_ppll}: $\sum_{j=1}^d\ppll^j \left( \boldsymbol{\Theta}; T \right)$.
    \item \emph{partially interval-censored dataset:} we use the interval-censored log-likelihood on the $E$ dimensions and the point-process log-likelihood on the $E^c$ dimensions (see \cref{eqn:pmbp_nll}): $\sum_{j=1}^e \npll^j \left( \boldsymbol{\Theta}; T \right) + \sum_{j=e+1}^d \ppll^j \left( \boldsymbol{\Theta}; T \right)$.
\end{itemize}

In what follows, we specify as \( \pmbp(d,e) \)-PP and \( \pmbp(d,e) \)-IC the \( \pmbp(d,e) \) model fit on the timestamp dataset and the partially interval-censored dataset, respectively. 
For brevity and whenever it is clear from context, we drop the dimensionalities \((d,e)\), and refer to the model fits as \( \pmbp \)-PP and \( \pmbp \)-IC. 
Also, for the \( \pmbp \)-IC fits, we specify $k$ -- the length of the observation window -- as \(\pmbp\)-IC[$k$].

{\bf{Inference:}}
\pioc{We infer a single $\boldsymbol{\Theta}$ for each group of sequences using a joint fitting procedure.
Since there are 50 groups in total, this process yields 50 estimates of $\boldsymbol{\Theta}$.
For each group of 50 sequences, we obtain $\boldsymbol{\Theta}$  that minimizes the joint negative log-likelihood, which is defined as $\mathcal{L}(\boldsymbol{\Theta}, T) = \sum_{i=1}^{50} \mathcal{L}_i(\boldsymbol{\Theta}, T),$
where $\mathcal{L}_i(\boldsymbol{\Theta}, T)$ represents the negative log-likelihood for the $i^{th}$ event sequence.}

{\bf{Metrics:}}
We evaluate parameter recovery error with four error metrics: the root-mean-squared error (RMSE) of each \(\pmbp\) parameter type $\{\hat{\boldsymbol{\alpha}}, \hat{\boldsymbol{\theta}}, \hat{\boldsymbol{\nu}}\}$ concerning the generating MHP parameters $\{\boldsymbol{\alpha}, \boldsymbol{\theta}, \boldsymbol{\nu} \}$ and the signed deviation $\Delta\rho = \rho(\hat{\boldsymbol{{\alpha}}}) - \rho(\boldsymbol{{\alpha}})$ of the spectral radius.

{\bf{Parameter Recovery Results:}}
\cref{fig:err} shows RMSE$(\boldsymbol{\alpha}$), RMSE$(\boldsymbol{\theta}$), RMSE$(\boldsymbol{\nu}$) and $\Delta\rho$ across model fits. \pioc{Each boxplot represents the distribution of errors calculated from 50 estimates of $\boldsymbol{\Theta}$, each obtained from a different group of 50 sequences in the synthetic dataset.} Within each subplot we have seven boxplots. The leftmost boxplot is the MHP fit, followed by the \pmbp-PP fit (i.e., the \pmbp fit on the timestamp dataset). The next five boxplots contain \pmbp-IC fits of increasingly wider observation windows 1, 2, 5, 10 and 20. Note that the MHP fit represents the case where we do not have either model mismatch and interval censoring error.

In each subplot of \cref{fig:err}, the gap between the first two boxplots (\ie MHP vs. \( \pmbp(2,1) \) fitted on timestamp data) indicates model mismatch error;
the gap between the second and third boxplots (\ie \( \pmbp(2,1) \) fitted on timestamp data vs. partially interval-censored data) indicates interval censoring error. 
The gaps between succeeding boxplots indicate the effect of wider observation windows.

\cref{fig:err} shows three conclusions.
First, model mismatch and interval censoring errors contribute to information loss relative to the MHP fit. 
Second, the approximation quality degrades as the observation window widens, indicating an increasing information loss of type (2).
Third, for parameters $\boldsymbol{\alpha}$ and $\boldsymbol{\nu}$, the model mismatch error appears negligible; it is only for higher values of the observation window length ($\geq 5$) that the performance starts degrading due to information loss error.
Both error types are present for $\boldsymbol{\theta}$.
See Sec. 14.1 of the SI for individual parameter fits.

Though we observe that the generating parameters are not always correctly recovered, we see in the rightmost subplot of \cref{fig:err} that, interestingly, the spectral radius estimation error $\Delta \rho$
is close to zero regardless of model mismatch and exhibits only slight underestimation for wide observation windows. 
This is particularly relevant, as $\rho(\boldsymbol{\alpha})$ is a meaningful quantity relating to information spread virality (for social media diffusions), disease infectiousness (for epidemiology), or local seismicity (in seismology). The result indicates that even when individual parameter fits are inaccurate, the MHP regime is correctly identified. 

{\bf{Behavior of $\Delta \rho$ in Higher Dimensions:}}
We further study the behavior of $\Delta \rho$ for varying MBP dimensions $e$ and model dimensionality $d$. 
We fix $T=100$ and $\rho(\boldsymbol{\alpha})=0.92$. 
Results for other error metrics are in Sec. 14.2 of the SI.

\begin{figure}[t!]
    \centering
    \includegraphics[width=\columnwidth]{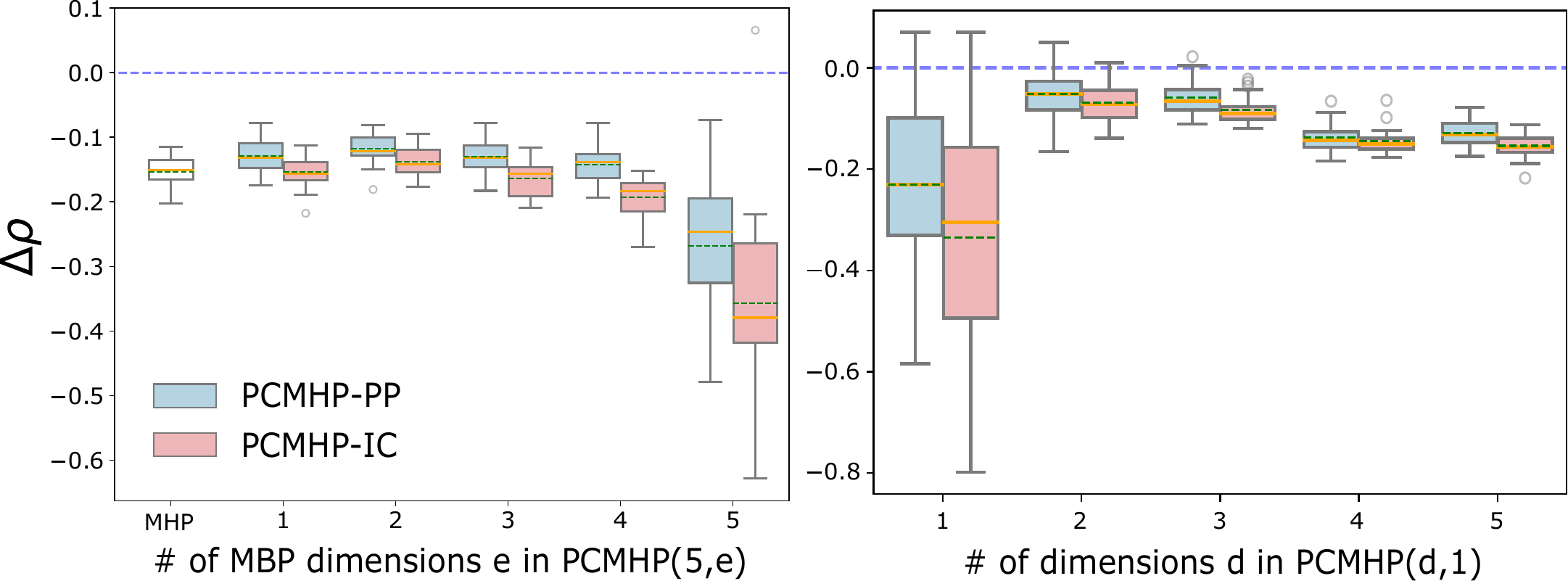}
    \caption{
         (Left) Relating the spectral radius estimation error $\Delta \rho$ of $\pmbp(5, e)$ and the number of MBP dimensions $e$. Note that $\pmbp(5, 0)$ is the MHP (\ie the data-generating process). (Right) Relating the spectral radius estimation error $\Delta \rho$ of $\pmbp(d, 1)$ and the model dimensionality $d$. In both plots, samples are drawn from a $d$-dimensional MHP with spectral radius $\rho(\boldsymbol{\alpha}) = 0.92$. Hyperparameters are $T=100$, $N_{sequences}=20$ and intervalsize=$1$. We fit two models for each $\pmbp$ column: $\pmbp-\text{PP}$ (\ie \(\pmbp\) fit on timestamp data on all dimensions) and $\pmbp-\text{IC}$ (\ie \(\pmbp\) fit on interval-censored data on the first $e$ dimensions and timestamp data on the last $d-e$ dimensions). The mean and median estimates are indicated by the dashed green lines and solid orange lines, respectively.
    }
    \label{fig:dimchange}
\end{figure}

In the left subplot of \cref{fig:dimchange}, we fix $d=5$ and observe how the spectral radius error $\Delta\rho$ varies with the number of MBP dimensions $e$.
Note that the leftmost boxplot represents the MHP fit (\ie, $e = 0$). 
Interestingly, we see that all \(\pmbp(5,e), e < 5\) flavors except the fully MBP case (\ie, $e=d=5$) can estimate the spectral radius as well as the MHP.
The gap between the estimated spectral radii and the generating value (blue dashed line) is attributable to the difficulty of recovering MHP parameters in higher dimensions.

In the right subplot of \cref{fig:dimchange}, we fix $e=1$ and observe how the spectral radius error $\Delta\rho$ varies with the dimensionality $d$ of the \(\pmbp(d,1)\). 
The recovery error is generally low (except for $d=1$). 
However, we see that the magnitude of the error $\Delta\rho$ increases with increasing dimensionality starting from $d=2$, which is not surprising since the number of parameters increases quadratically as we increase the dimensionality of the process. 
We also see that fitting with a fully MBP model ($d=1$) does not show good recovery performance due to information loss, implying the necessity of having at least one cross-exciting dimension (\ie, $d-e \neq 0$).

% !TEX root=../main.tex
%
\section{YouTube Popularity Prediction}
\label{subsection:popularityprediction}
In this section, we evaluate \( \pmbp(d,e) \)'s performance in predicting the popularity of YouTube videos. 
For each video, we capture information about three dimensions -- \emph{views}, external \emph{shares} and \emph{tweets} linking to the videos -- over the time period $[0, T^{train})$.
We measure time in days relative to the time of posting on Youtube. 
The first two dimensions (the views and shares) are observed as daily counts, \ie, $E = \{views, shares\}$. 
The third dimension (tweets) is provided as event times, \ie $E^c =\{tweets\}$.
Given this data setup, we use \( \pmbp(3,2) \) to predict the daily counts of views and shares and the timestamp of the tweets posted over the period $[T^{train}, T^{test})$.

{\bf{Interval-Censored Forecasting with $\pmbp$:}}
%% MAR: describe first the setup
To each YouTube video corresponds a partially interval-censored Hawkes realization.
A straightforward approach to predict the unfolding of the realization during $[T^{train}, T^{test})$ is to sample timestamps from \( \pmbp(3,2) \) on each of the three dimensions, conditioned on data before $T^{train}$; we then interval-censor the first two dimensions.
In practice, sampling individual views takes considerable computational effort due to their high background rates, sometimes in the order of millions of views per day, and usually at least an order of magnitude larger than shares and tweets.

\begin{figure}[!htbp]
    \centering
    \includegraphics[width=\columnwidth]{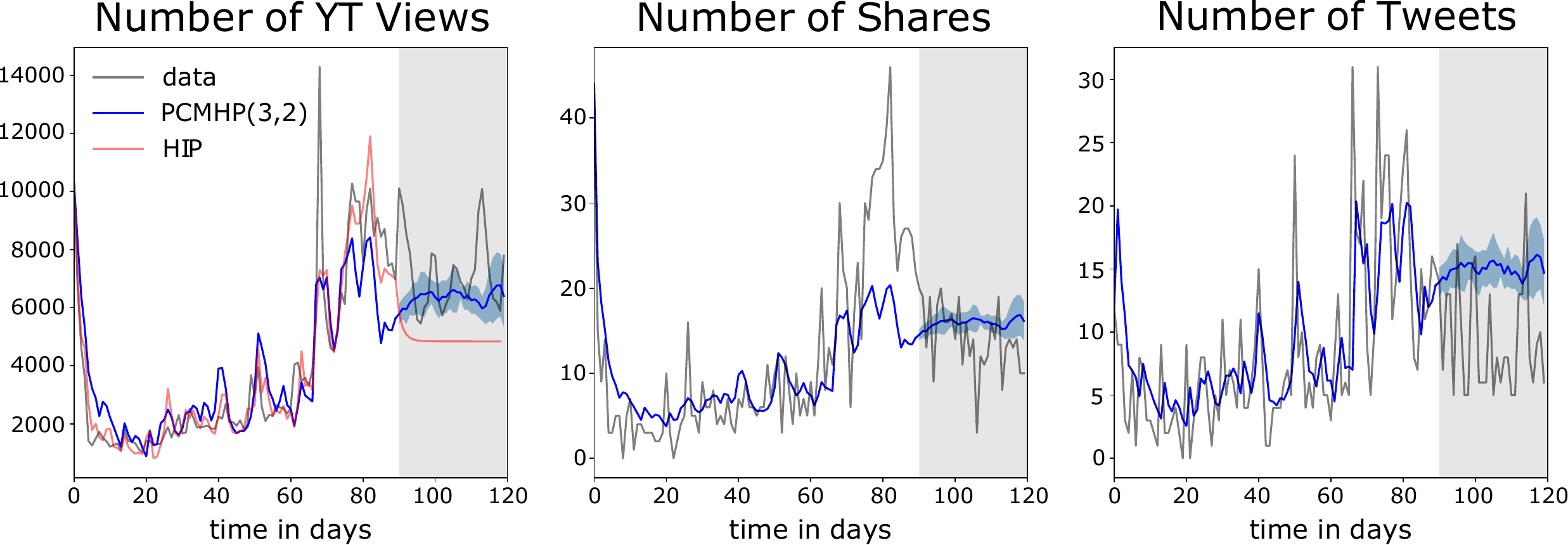}
    \caption{
        Comparison of fits and predictions of our proposal \( \pmbp(3,2) \) and the baseline HIP \cite{Rizoiu2017} for views (left), shares (center) and tweets (right) for a sample video from \actived: a trailer for the 2014 movie Whiplash (id $7d\_jQycdQGo$).
        The first 90 days are used to fit model parameters, while the next 30 days (indicated by the \textcolor{gray}{gray shaded area}) are unseen by the model and used for evaluation.
        HIP does not predict the share and tweet counts, as it treats these as exogenous inputs. The \textcolor{blue}{blue shaded area} shows prediction uncertainty computed for the $\pmbp(3,2)$ fits.}
    \label{fig:popularity-sample-12}
\end{figure}

Below is an efficient procedure to calculate expected counts that leverages the compensator $\boldsymbol{\Xi}_E$ and requires sampling only the $E^c$ dimensions (\ie, tweets). Let $\mathcal{P}[T^{train}, T^{test}) = \bigcup_{i=1}^{P-1} [o_i, o_{i+1}),$ where $o_1 = T^{train}$ and $o_{P} = T^{test}$, be a partition of $[T^{train}, T^{test})$.

\begin{enumerate}
    \item Sample only the $E^c$ dimensions on $[T^{train}, T^{test})$.
    \item Compute expected counts on $\mathcal{P}[T^{train}, T^{test})$ as $\{\boldsymbol{\Xi}_E(o_{i+1})$ - $\boldsymbol{\Xi}_E(o_{i}) \vert i \in 1 \cdots P-1\}$.
    \item Compute the average of $\{\boldsymbol{\Xi}_E(o_{i+1})$ - $\boldsymbol{\Xi}_E(o_{i}) \vert i \in 1 \cdots P-1\}$ across samples.
\end{enumerate}

More details of this scheme and a comparison with the standard method of sampling both $E^c$ and $E$ dimensions are provided in Sec. 13 of the SI \cite{appendix}.

{\bf{Dataset, Experimental Setup and Evaluation:}}
We use two subsets of the \actived dataset~\cite{Rizoiu2017} for model fitting and evaluation. 
The first subset -- dubbed \actived $20\%$ -- contains a $20\%$ random sample of the \actived dataset~\cite{Rizoiu2017}, \ie, $2,834$ videos published between 2014-05-29 and 2014-12-26. 
The second subset -- dubbed \emph{\textsc{dynamic videos}} -- contains videos with which users engage significantly during the test period.
It is known that users' attention to YouTube videos decays with time~\cite{Crane2008, Wu2019}; therefore, the daily views of most \actived videos hover around zero more than 90 days after their upload.
We select the 585 dynamic videos with the standard deviations of the views, tweets, and shares counts on days $21-90$ higher than the median values on each of the three measures.
Technical details of the filtering are in Sec. 15.1 of the SI \cite{appendix}. 

For each video, we tune $\pmbp$ hyperparameters and parameters using the first $90$ days of daily view counts, share count to external platforms, and the timestamps of tweets that mention each video ($T^{train} = 90$). It is known that generative models are suboptimal for prediction \cite{Mishra2016} and have to be adapted to the prediction task for better performance. Similar to HIP, we implement dimension weighting and parameter regularization in the likelihood. Full technical details of the fitting procedure are provided in Sec. 15.2 of the SI.

The days $91-120$ are used for evaluation ($T^{test} = 120$). 
We measure prediction performance using the Absolute Percentile Error (APE) metric~\cite{Rizoiu2017}, which accounts for the long-tailness of online popularity -- \eg, the impact of an error of 10,000 views is very different for a video getting 20,000 views per day compared to a video getting 2 million views a day.
We first compute the percentile scale of the number of views accumulated between days 91 and 120.
APE is defined as:
\begin{equation*}
    APE = | \text{Per}(\hat{N}_{120}) - \text{Per}(N_{120})|
\end{equation*}
where $\hat{N}_{120}$ and $N_{120}$ are the predicted and observed number of views between days $91$ and $120$; the function $\text{Per}(\cdot)$ returns the percentile of the argument on the popularity scale.

{\bf{Models and Baseline:}} 
We consider two $3$-dimensional \pmbp models: \( \pmbp(3,2) \) and \( \pmbp(3,3) \).
The former treats the tweets as a Hawkes dimension (see \cref{def:pmbp_definition}) and is thus susceptible to computational explosion for high tweet counts given the quadratic complexity of computing cross- and self-excitation.
The latter is an inhomogeneous Poisson process with no self- or cross-exciting dimension. 
We, therefore,  fit \( \pmbp(3,2) \) solely on videos that have less than 1000 tweets on days $1-90$; we fit \( \pmbp(3,3) \) on all videos.

We use as a baseline the Hawkes Intensity Process (HIP)\alexs{~\cite{Rizoiu2017}}, 
%the current parametric state-of-the-art in 
\pioc{a parametric popularity prediction model} discussed in \cref{section:background}. 
HIP, however, is designed for use in a forecasting setup. 
That is, HIP requires the actual counts of tweets and shares in the prediction window $[T^{train}, T^{test})$ to get forecasts for the view counts on $[T^{train}, T^{test})$. 
To adapt HIP for the prediction setup (\ie, the tweets and shares are not available at test time),
we feed HIP for each of the days 91-120 the time-weighted average of the daily tweet and share counts on $1-90$, \ie $\frac{1}{\sum_{t=1}^{90} t} \sum_{t=1}^{90} t \cdot \#\text{tweets}(t)$ and $\frac{1}{\sum_{t=1}^{90} t} \sum_{t=1}^{90} t \cdot \#\text{shares}(t)$,
which assigns a higher weight to more recent counts.

%!TEX root=../main.tex
%

\begin{table}[tbp]
    \centering
    \caption{
        Performance comparison of \( \pmbp(3,3) \), \( \pmbp(3,2) \) and HIP on (a) a random sample that comprises $20\%$  of the videos in \actived, and (b) the set of dynamic videos from \actived: mean, median, and standard deviation of the percentile errors for each model. Best-performing score in bold.
    }
    \label{tab:ytperformance}
    \begin{tabular}{lccc|ccc}
           & \multicolumn{3}{c}{ACTIVE20\% (n=2834)}   & \multicolumn{3}{c}{DYNAMIC (n=585)}                                                       \\ \hline
           &  \makecell{\pmbp \\ (3,3)}     & \makecell{\pmbp \\ (3,2)} & HIP  & \makecell{\pmbp \\ (3,3)} & \makecell{\pmbp \\ (3,2)}   & HIP                           \\ \hline
    Mean   & \textbf{4.82} & 7.36      & 8.12 & 10.86     & \textbf{7.28}                         & 9.31                          \\
    Median & \textbf{2.55} & 4.69      & 4.96 & 4.82      & \cellcolor[HTML]{FFFFFF}\textbf{3.79} & \cellcolor[HTML]{FFFFFF}4.73  \\
    StdDev    & \textbf{7.13} & 8.34      & 9.89 & 14.24     & \cellcolor[HTML]{FFFFFF}\textbf{9.58} & \cellcolor[HTML]{FFFFFF}11.89 \\ \hline
\end{tabular}
\end{table}

{\bf{Results:}} \cref{fig:popularity-sample-12} illustrates the fits of \( \pmbp(3,2) \) and the baseline HIP \cite{Rizoiu2017}  for a sample video from \actived. 
Visibly, we see that \( \pmbp(3,2)\) and HIP have comparable fits of the popularity dynamics (left column) during the training period (unshaded area), but \( \pmbp(3,2) \) outputs a much tighter fit during the test period (gray shaded area).
We also observe two advantages of \pmbp.
First, being a multivariate process that captures endogenous dynamics across its dimensions, \( \pmbp(3,2) \) provides a prediction for future share and tweet counts (center and left columns), in addition to the number of views.
In contrast, HIP treats views (i.e. popularity) as exogenously driven by tweets and shares and thus can only predict the views' dimension. 
Second, \pmbp can quantify the uncertainty of the popularity prediction by sampling multiple unfoldings of a realization and computing the variance of the samples (shown as the blue shaded area in \cref{fig:popularity-sample-12}).

In \cref{tab:ytperformance}, we tabulate the mean, median and standard deviation of percentile errors for \( \pmbp(3,3) \), \( \pmbp(3,2) \), and HIP on \actived $20\%$ and \textsc{dynamic videos}.
We observe that the \pmbp flavors consistently outperform the baseline HIP on both datasets.
Visibly, on \actived $20\%$, \( \pmbp(3,3) \) outperforms \( \pmbp(3,2) \). 
This is because most videos in \actived $20\%$ do not exhibit much activity during the test period.
Consequently, as a nonhomogeneous Poisson process with no self-excitation, \( \pmbp(3,3) \) fits better such flat trends than the self-exciting \( \pmbp(3,2) \) and HIP models.
On \textsc{dynamic videos}
we see a reversal of performance ranking: \( \pmbp(3,2) \) performs best, followed by HIP and \( \pmbp(3,3) \). 
This result corroborates our claim in \cref{subsection:discussion}  that applying the heuristic of censoring event times leads to information loss. We see that \( \pmbp(3,2) \) (trained on tweet times) can better capture the popularity dynamics of the most complex videos (which are also the most interesting) compared to \( \pmbp(3,3) \) (trained on tweet counts).
% !TEX root=../main.tex
%
\section{Interaction Between COVID-19 Cases and News}
\label{subsection:covid}
In the previous section we have validated the predictive power of the \pmbp. Here, we shift our attention to the interpretability of \pmbp-fitted parameters.
We showcase how \pmbp can link online and offline streams of events by learning the interaction between the COVID-19 daily case counts and publication dates of COVID-19-related news articles for 11 different countries during the early stage of the pandemic.

\begin{figure}[t!]
    \centering
    \includegraphics[width=\columnwidth]{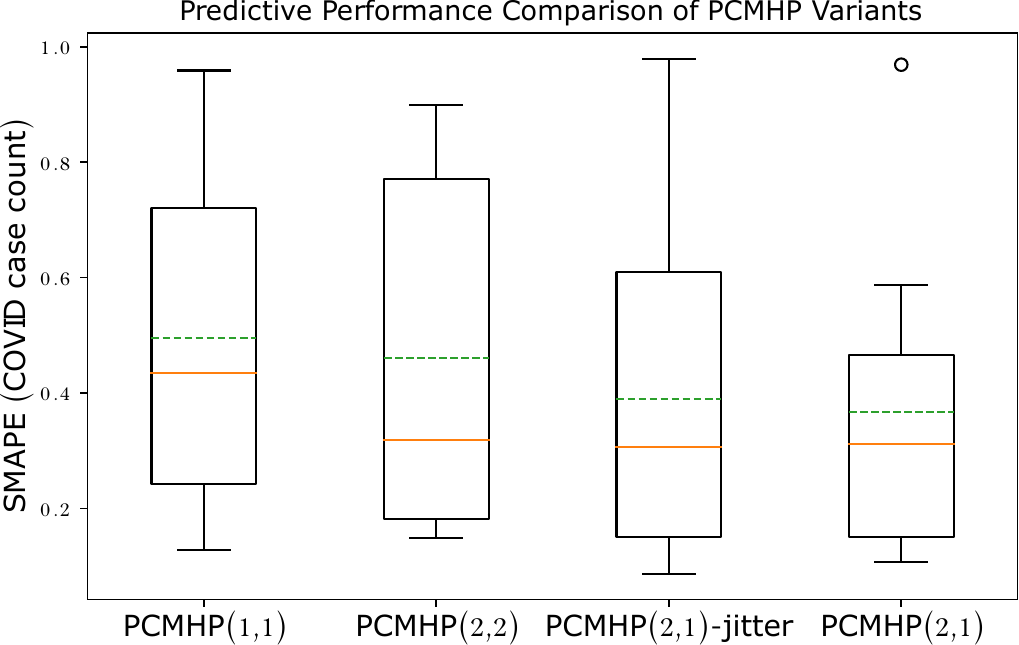}
    \caption{
            Performance comparison of \( \pmbp(1,1) \), \( \pmbp(2,2) \), \( \pmbp(2,1)\)-jitter and \( \pmbp(2,1) \) on the COVID case count prediction task over our sample of 11 countries.
            The dashed line and solid line indicate the mean and median
            estimates, respectively.
        }
    \label{fig:covid-prediction-boxplot}
\end{figure}

{\bf{Dataset:}}
We curate and align two data sources.

The first dataset contains COVID daily case counts from the Johns Hopkins University \cite{Dong2020}. 
The dataset is a set of date-indexed spreadsheets containing COVID reported case counts split by country and region. 
We focus on the following 11 countries: UK, USA, Brazil, China, France, Germany, India, Italy, Spain, Sweden, and the Philippines. 
We select the same countries as
\cite{Browning2021}, to which we add the Philippines.

The second dataset contains timestamps of COVID-19-related news articles provided by the NLP startup Aylien \cite{Aylien2020}. 
This dataset is a dump of COVID-related English news articles from 440 major sources from November 2019 to July 2020.
We filter the Aylien dataset for news articles that mention the selected 11 countries in the headline. 
To improve relevancy, for China, we also use several COVID-related keywords (such as \textit{coronavirus}, \textit{covid} and \textit{virus}).
Lastly, we only select articles from popular news sites with an Alexa rank of less than 150.
Such news sources include Google News and Yahoo! News.

For each country, we fit \( \pmbp(2,1) \) with $E=\{cases\}$ and $E^c=\{news\}$. We consider as $t=0$ the first day on which a minimum of 10 cases were recorded.
Except for China, which had cases as early as January 2020, the initial time for each country in our sample lies between February and March 2020.
We only consider data until $t = 120$, with time measured in days.

{\bf{Incorporating News Information:}} 
To demonstrate the utility of news information in modeling COVID case counts, we compare the predictive performance of \(\pmbp(2,1)\) with three variants that leverage different granularities of news information. First, we compare with \(\pmbp(1,1)\) which does not use news information at all. Second, we compare with \(\pmbp(2,2)\) that uses daily aggregated news counts. Lastly, to test whether exact timing of news is important, we disaggregate daily news counts by adding a uniform jitter to each time, similar to what is done in \cite{Unwin2021}, and fit \(\pmbp(2,1)\) to this dataset. We call this baseline \(\pmbp(2,1)\)-jitter. 

\begin{figure}[t!]
    \includegraphics[width=\columnwidth]{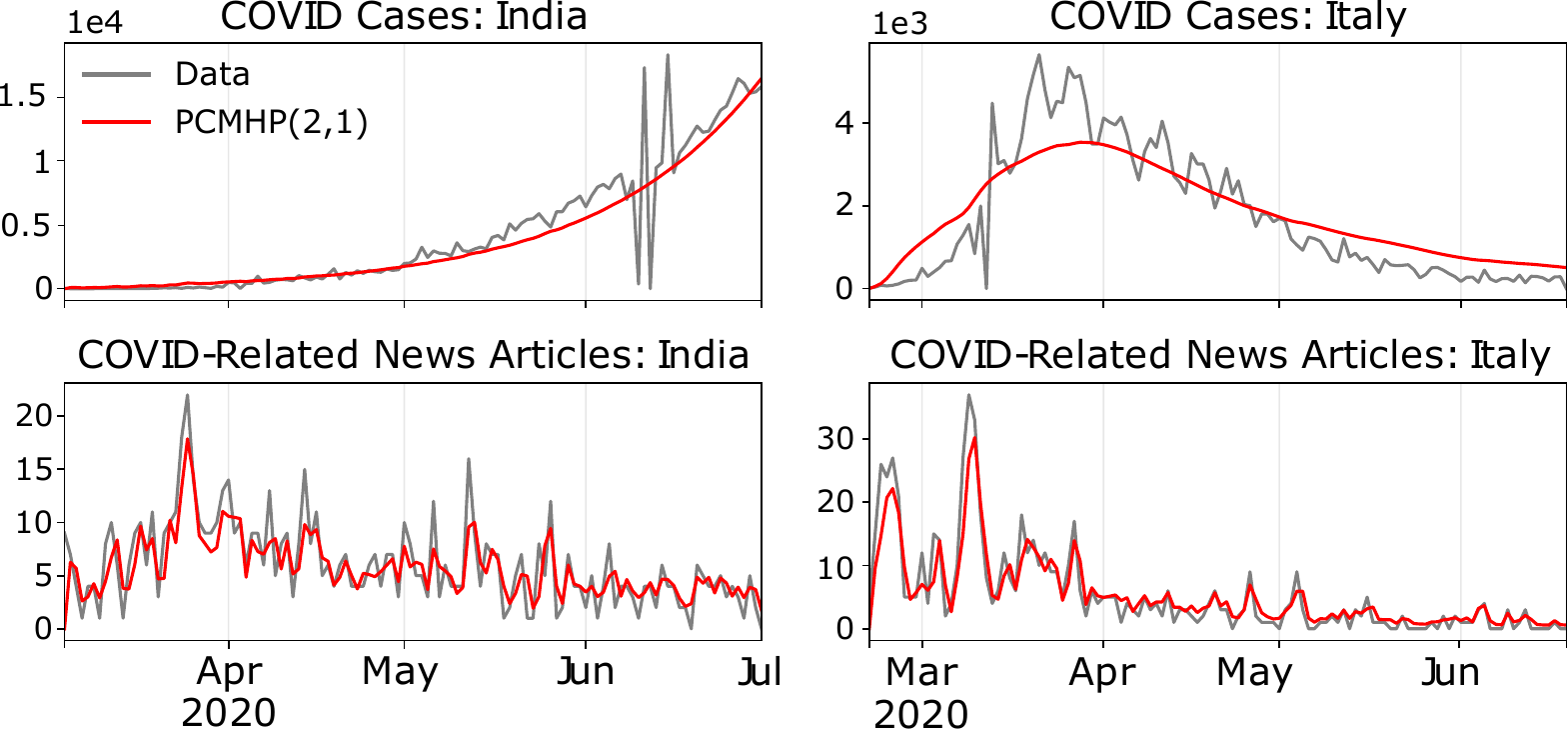}
    \caption{
        Observed and \( \pmbp(2,1) \)-fitted daily COVID-19 case counts (top row) and COVID-19-related news articles (bottom row) for India (left column) and Italy (right) during the early stage of the outbreak.
    }
    \label{fig:covid-fits-india-italy}
\end{figure}

Similar to \cref{subsection:popularityprediction}, we split our timeframe into a training period $[0, T^{train}=90)$ and a testing period $[T^{train}, T^{test}=120)$.
In our training period, we fit the models and perform hyperparameter tuning; in our testing period, we sample from the fitted models and evaluate performance. 
We measure performance using the Symmetric Mean Absolute Percentage Error (SMAPE), given by $SMAPE = \frac{1}{n} \sum_{t=1}^n \frac{|F_t - A_t|}{|A_t| + |F_t|},$ where $F_t$ and $A_t$ are the forecasted and actual values at time $t$, respectively.

Across our sample of 11 countries, we see in \cref{fig:covid-prediction-boxplot} that \(\pmbp(2,1)\) has the best performance compared to the three baselines and incorporating more granular news information leads to better predictive performance.
We observe that the news-agnostic \(\pmbp(1,1)\) and the day aggregated \(\pmbp(2,2)\) models do not fit the data well and cannot capture the complex COVID case count dynamics. This supports our claim in \cref{subsection:discussion} that application of data-altering heuristics leads to loss of information. 
However, by incorporating timestamped news information, we see significant performance improvement and we can match the trend in the case time series. We also see subtle performance improvement by incorporating exact news times (\(\pmbp(2,1)\)-jitter vs. \(\pmbp(2,1)\)).

%!TEX root=../main.tex
%
\begin{table}[tbp]
    \centering
    \caption{ $K$-means cluster centroids on the parameters obtained by fitting $\pmbp(2,1)$ on the case count and news article dataset.}
    \label{tab:covid-country-clusters}
    \setlength{\tabcolsep}{2.1pt}
    \begin{tabular}{lllllllllll}
        \toprule
        Cluster & $\theta^{11}$ & $\theta^{12}$ & $\theta^{21}$ & $\theta^{22}$ & $\alpha^{11}$ & $\alpha^{12}$ & $\alpha^{21}$ & $\alpha^{22}$ & $\nu^1$ & $\nu^2$ \\
        \midrule
        UK, German, Spain & 0.76 & 0.12 & 1.82 & 1.84 & 0.79 & 3.60 & 0.02 & 0.42 & 0.03 & 0.28\\
        Brazil & 0.13 & 0.01 & 1.89 & 2.46 & 1.08 & 5.48 & 0 & 0.38 & 0 & 1.47\\
        China, France & 0.62 & 4 & 1.70 & 3.55 & 0.68 & 0.73 & 0.3 & 0.39 & 0 & 0.05\\
        US, Italy, Sweden & 0.67 & 0.22 & 1.61 & 1.51 & 0.93 & 0.73 & 0.006 & 0.65 & 0.29 & 0.59\\
        India, Philippines & 0.12 & 2.28 & 2.15 & 1.88 & 1.35 & 0.54 & 0.007 & 0.58 & 0.08 & 0.66\\
        \bottomrule
    \end{tabular}
\end{table}

{\bf{Results:}} \cref{fig:covid-fits-india-italy} shows the daily COVID-19 case counts and daily news article volume of the \( \pmbp(2,1) \) fits for India and Italy. We show the plots for the other countries, the table of parameter estimates, and the goodness-of-fit analysis in Sec. 16 of the SI \cite{appendix}.
Visible from \cref{fig:covid-fits-india-italy}, \( \pmbp(2,1) \) captures well the dynamics of both countries. 
Based on the sample-based fit score introduced in the SI, the actual COVID-19 case counts for India and Italy fall within the model's prediction interval for $97\%$ and $61\%$ of the time, respectively.  

{\bf{Cluster countries based on model fittings:}}
The parameters capture different aspects of the interaction between news and cases.
Here, we cluster the fitted parameter sets across countries to identify groups that have similar diffusion profiles. 
To render the scale of parameters comparable across countries, we rescale the maximum daily number of cases for each country over the considered timeframe to be 100, fit the \( \pmbp(2,1) \) on this scaled data, and perform $K$-means clustering on the resulting parameter sets.

\begin{figure}[tbp]
    \centering
    \includegraphics[width=\columnwidth]{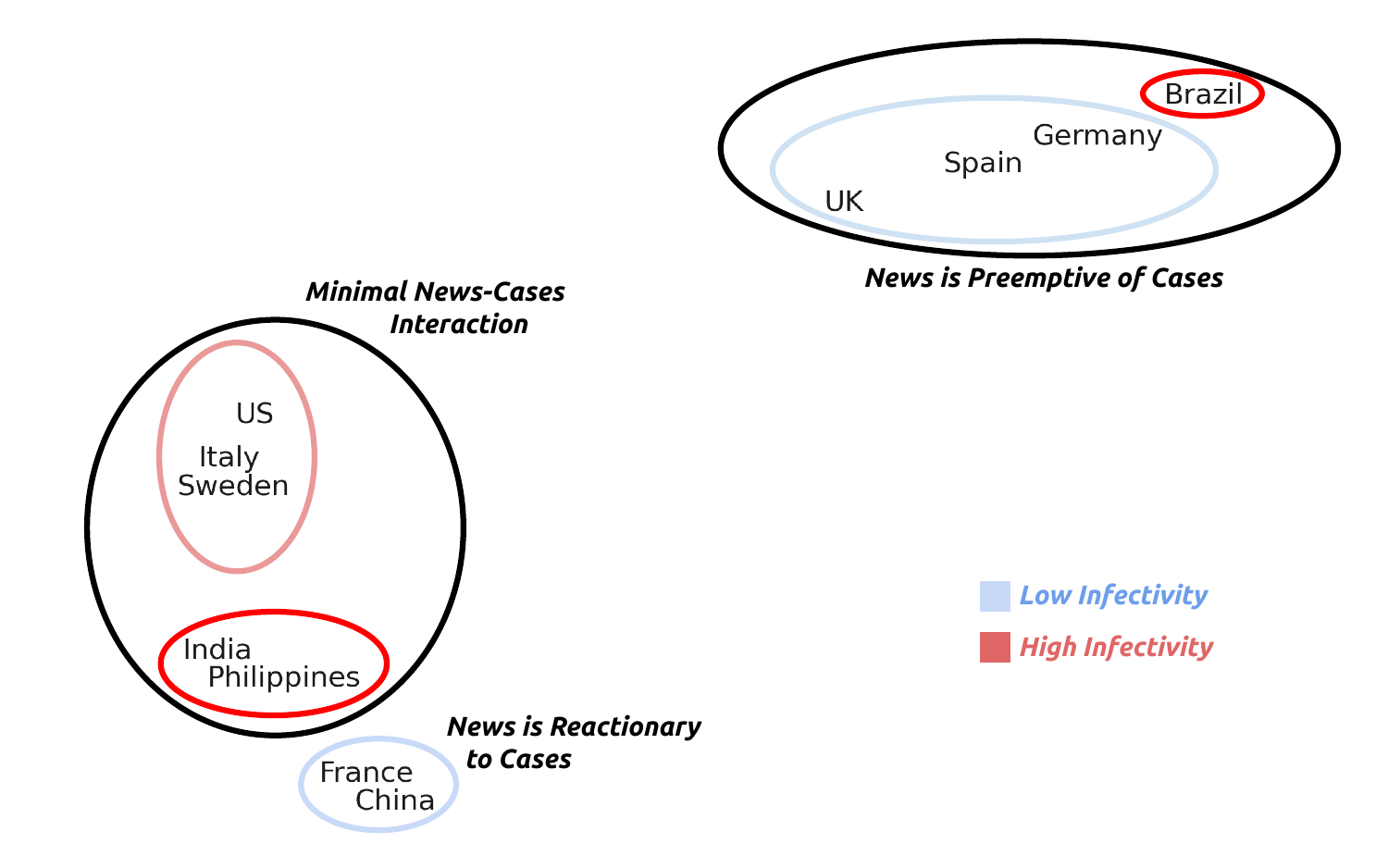}
    \caption{Labeled tSNE visualization of the clusters obtained from the fitted \( \pmbp(2,1) \) parameters across the 11 countries we consider.}
    \label{fig:covid-cluster}
\end{figure}

The $k=5$ clusters are shown in \cref{tab:covid-country-clusters} and visualized in \cref{fig:covid-cluster} using t-SNE \cite{vdMaaten2008}.
The first cluster (the UK, Germany, Spain) has high $\alpha^{12}$ and low $\alpha^{11}$. 
The second cluster -- made solely of Brazil --  has both a high $\alpha^{12}$ and a very high $\alpha^{11}$. 
With high $\alpha^{12}$, the two clusters contain countries where \textit{news strongly preempts cases}. 
The third cluster (China and France) has a high $\alpha^{21}$ indicative of \textit{news playing a reactive role to cases}. 
The fourth cluster (US, Italy, Sweden) and fifth cluster (India, Philippines) both have low $\alpha^{12}$ and $\alpha^{21}$, indicating \textit{little interaction between news and cases}. 
We notice that COVID infectiousness is much higher in the fifth cluster (India, Philippines), with $\alpha^{11}$ greater than one (each case generates more than one case) and $\theta^{11}$ lowest across all clusters (slow decay, therefore long influence from cases to cases).
Our fits indicate that India and the Philippines are countries particularly affected by COVID-19 in the early days.

% !TEX root=../main.tex
%
\section{Summary and Future Work}
\label{subsection:summary}

This work introduces the Partially Censored Multivariate Hawkes Process (\pmbp), a generalization of the MHP where we take the conditional expectation of a subset of dimensions over the stochastic history of the process.
The \pmbp is motivated by the fact that the MHP cannot directly be fit to partially interval-censored data; the \pmbp can be used to approximate MHP parameters via a correspondence of parameters.

In this paper, we derive the conditional intensity function of the \pmbp by considering the impulse response to the associated LTI system. Additionally, we derive its regularity conditions which leads to a subcritical process; which we find generalizes regularity conditions of the multivariate Hawkes process and the previously proposed MBP process. 
The MLE loss function is also derived for the partially interval-censored setting. To test the practicality of our proposed approach, we consider three empirical experiments.

First, we test the capability of the \pmbp in recovering multivariate Hawkes process parameters in the partially interval-censored setting. By using synthetic data, we investigate the information loss from model mismatch and the interval-censoring of the timestamped data. Our results show that the fitted \pmbp can approximate the parameters and recover the spectral radius of the original multivariate Hawkes process used to generate the data. 

Second, we demonstrate the predictive capability of the \pmbp model by applying it to YouTube popularity prediction and showing that it outperforms \pioc{the popularity estimation algorithm Hawkes Intensity Process \cite{Rizoiu2017}}. 

Third, to demonstrate interpretability of the \pmbp parameters, we fit the process to a curated dataset of COVID-19 cases and COVID-19-related news articles during the early stage of the outbreak in a sample of countries. By inspecting the country-level parameters, we show that there is a demonstrable clustering of countries based on how news predominantly played its role: whether it was reactionary, preemptive, or neutral to the rising level of cases.

{\bf{Limitations and Future Work:}} The number of parameters estimated with the \( \pmbp(d,e) \) increases quadratically with the process dimension $d$, meaning the curse of dimensionality applies. 
However, since the primary use case for \( \pmbp(d,e) \) is in popularity prediction across social media platforms (i.e., where the dimensions represent platforms), which is a low-dimensional scenario, the curse of dimensionality is not a significant concern. 
Difficulties would arise, however, if \( \pmbp(d,e) \) were applied at the user level (i.e., where the dimensions correspond to individual users). 
In such cases, future work would be required to extend PCMHP by employing scalable variants of the Hawkes process. 
These approaches generally reduce the $d$ dimensions into a more compact $r$-dimensional space ($r \ll d$) through factorization of the interaction matrix \cite{lemonnier2017multivariate,nickel2020learning}. 
This is a direction for future research.

There are three other areas where future work can be explored. First, theoretical analysis on the approximation error of the model mismatch (\ie, fitting Hawkes data to the \pmbp model) should be performed, since we only performed an empirical evaluation in this work. Second, methods to approximate the conditional intensity should be investigated, as the current solution relies on the computationally heavy discrete convolution approximation. Lastly, given that the \( \pmbp(d,e) \), by construction, is not self- and cross-exciting in the $E$ dimensions, an open research question is whether we can construct a process that retains the self- and cross-exciting properties in all dimensions whilst also being flexible enough to be used in the partially interval-censored setting.

\section{Acknowledgments}
\noindent The work was partially supported by the National Science Centre, Poland, under Grant 2021/41/B/HS6/02798.

\bibliographystyle{IEEEtran}
\bibliography{references}

\begin{thebibliography}{10}
\providecommand{\url}[1]{#1}
\csname url@samestyle\endcsname
\providecommand{\newblock}{\relax}
\providecommand{\bibinfo}[2]{#2}
\providecommand{\BIBentrySTDinterwordspacing}{\spaceskip=0pt\relax}
\providecommand{\BIBentryALTinterwordstretchfactor}{4}
\providecommand{\BIBentryALTinterwordspacing}{\spaceskip=\fontdimen2\font plus
\BIBentryALTinterwordstretchfactor\fontdimen3\font minus
  \fontdimen4\font\relax}
\providecommand{\BIBforeignlanguage}[2]{{%
\expandafter\ifx\csname l@#1\endcsname\relax
\typeout{** WARNING: IEEEtran.bst: No hyphenation pattern has been}%
\typeout{** loaded for the language `#1'. Using the pattern for}%
\typeout{** the default language instead.}%
\else
\language=\csname l@#1\endcsname
\fi
#2}}
\providecommand{\BIBdecl}{\relax}
\BIBdecl

\bibitem{Rizoiu2017}
M.~A. Rizoiu, L.~Xie, S.~Sanner, M.~Cebrian, H.~Yu, and P.~Van~Henteryck,
  ``Expecting to be {HIP}: {Hawkes} intensity processes for social media
  popularity,'' \emph{WWW 2017}, pp. 735--744, 2017.

\bibitem{Hawkes1971}
A.~G. Hawkes, ``\BIBforeignlanguage{en}{Spectra of some self-exciting and
  mutually exciting point processes},''
  \emph{\BIBforeignlanguage{en}{Biometrika}}, vol.~58, no.~1, pp. 83--90, 1971.

\bibitem{Rizoiu2017T}
M.-A. Rizoiu, Y.~Lee, S.~Mishra, and L.~Xie, \emph{A Tutorial on Hawkes
  Processes for Events in Social Media}.\hskip 1em plus 0.5em minus 0.4em\relax
  Frontiers of Multimedia Research, 12 2017, pp. 191--218.

\bibitem{Browning2021}
R.~Browning, D.~Sulem, K.~Mengersen, V.~Rivoirard, and J.~Rousseau, ``Simple
  discrete-time self-exciting models can describe complex dynamic processes: A
  case study of covid,'' \emph{PLoS ONE}, pp. 1--28, 2021.

\bibitem{Shlomovich2020}
L.~Shlomovich, E.~A. Cohen, N.~Adams, and L.~Patel, ``Parameter estimation of
  binned hawkes processes,'' \emph{Journal of Computational and Graphical
  Statistics}, vol.~31, no.~4, pp. 990--1000, 2022.

\bibitem{Daley2003}
D.~Daley and D.~Vere-Jones, \emph{An introduction to the theory of point
  processes. {V}ol. {I}}, 2nd~ed., ser. Probability and its Applications (New
  York).\hskip 1em plus 0.5em minus 0.4em\relax New York: Springer-Verlag,
  2003.

\bibitem{Rizoiu2022}
M.-A. Rizoiu, A.~Soen, S.~Li, P.~Calderon, L.~Dong, A.~K. Menon, and L.~Xie,
  ``{Interval-censored Hawkes processes},'' \emph{Journal of Machine Learning
  Research}, vol.~23, no. 338, pp. 1--84, 2022.

\bibitem{Unwin2021}
H.~J.~T. Unwin, I.~Routledge, S.~Flaxman, M.-A. Rizoiu, S.~Lai, J.~Cohen, D.~J.
  Weiss, S.~Mishra, and S.~Bhatt, ``\BIBforeignlanguage{en}{Using {Hawkes}
  {Processes} to model imported and local malaria cases in near-elimination
  settings},'' \emph{\BIBforeignlanguage{en}{PLOS Computational Biology}},
  vol.~17, no.~4, Apr. 2021.

\bibitem{Zhao2015}
Q.~Zhao, M.~A. Erdogdu, H.~Y. He, A.~Rajaraman, and J.~Leskovec, ``Seismic: A
  self-exciting point process model for predicting tweet popularity,'' in
  \emph{SIGKDD 2015}, 2015, pp. 1513--1522.

\bibitem{Kobayashi2016}
R.~Kobayashi and R.~Lambiotte, ``Tideh: Time-dependent hawkes process for
  predicting retweet dynamics,'' in \emph{ICWSM 2016}, pp. 191--200.

\bibitem{Mishra2016}
S.~Mishra, M.~A. Rizoiu, and L.~Xie, ``Feature driven and point process
  approaches for popularity prediction,'' \emph{CIKM 2016}, pp. 1069--1078.

\bibitem{Zadeh2022}
A.~Zadeh and R.~Sharda, ``How can our tweets go viral? point-process modelling
  of brand content,'' \emph{Information \& Management}, vol.~59, no.~2, p.
  103594, 2022.

\bibitem{Crane2008}
R.~Crane and D.~Sornette, ``Robust dynamic classes revealed by measuring the
  response function of a social system,'' \emph{PNAS}, vol. 105, no.~41, pp.
  15\,649--15\,653, 2008.

\bibitem{Krohn2019}
R.~Krohn and T.~Weninger, ``Modelling online comment threads from their
  start,'' in \emph{2019 IEEE International Conference on Big Data}, 2019, pp.
  820--829.

\bibitem{yan2016}
Q.~Yan, S.~Tang, S.~Gabriele, and J.~Wu, ``\BIBforeignlanguage{en}{Media
  coverage and hospital notifications: {Correlation} analysis and optimal media
  impact duration to manage a pandemic},''
  \emph{\BIBforeignlanguage{en}{Journal of Theoretical Biology}}, vol. 390, pp.
  1--13, Feb. 2016.

\bibitem{chunara2012}
R.~Chunara, J.~R. Andrews, and J.~S. Brownstein,
  ``\BIBforeignlanguage{en}{Social and {News} {Media} {Enable} {Estimation} of
  {Epidemiological} {Patterns} {Early} in the 2010 {Haitian} {Cholera}
  {Outbreak}},'' \emph{\BIBforeignlanguage{en}{The American Journal of Tropical
  Medicine and Hygiene}}, vol.~86, no.~1, pp. 39--45, Jan. 2012.

\bibitem{yan2020}
Q.~Yan, Y.~Tang, D.~Yan, J.~Wang, L.~Yang, X.~Yang, and S.~Tang,
  ``\BIBforeignlanguage{en}{Impact of media reports on the early spread of
  {COVID}-19 epidemic},'' \emph{\BIBforeignlanguage{en}{Journal of Theoretical
  Biology}}, vol. 502, p. 110385, Oct. 2020.

\bibitem{Kirchner2016}
M.~Kirchner, ``\BIBforeignlanguage{en}{Hawkes and {INAR}( $\infty$ )
  processes},'' \emph{\BIBforeignlanguage{en}{Stochastic Processes and their
  Applications}}, vol. 126, no.~8, pp. 2494--2525, Aug. 2016.

\bibitem{Kirchner2017}
------, ``\BIBforeignlanguage{en}{An estimation procedure for the {Hawkes}
  process},'' \emph{\BIBforeignlanguage{en}{Quantitative Finance}}, vol.~17,
  no.~4, pp. 571--595, Apr. 2017.

\bibitem{Cheysson2020}
F.~Cheysson and G.~Lang, ``{Spectral estimation of Hawkes processes from count
  data},'' \emph{The Annals of Statistics}, vol.~50, no.~3, pp. 1722 -- 1746,
  2022.

\bibitem{Shlomovich2021}
L.~Shlomovich, E.~A.~K. Cohen, and N.~Adams, ``A parameter estimation method
  for multivariate binned {{Hawkes}} processes,'' \emph{Statistics and
  Computing}, vol.~32, no.~6, p.~98, Oct. 2022.

\bibitem{Schneider2023}
P.~J. Schneider and T.~A. Weber, ``\BIBforeignlanguage{en}{Estimation of
  self-exciting point processes from time-censored data},''
  \emph{\BIBforeignlanguage{en}{Physical Review E}}, vol. 108, no.~1, p.
  015303, Jul. 2023.

\bibitem{Bertozzi2020}
A.~L. Bertozzi, E.~Franco, G.~Mohler, M.~B. Short, and D.~Sledge,
  ``\BIBforeignlanguage{en}{The challenges of modeling and forecasting the
  spread of {COVID}-19},'' \emph{\BIBforeignlanguage{en}{PNAS}}, vol. 117,
  no.~29, pp. 16\,732--16\,738, Jul. 2020.

\bibitem{Chiang2020}
W.-H. Chiang, X.~Liu, and G.~Mohler, ``\BIBforeignlanguage{en}{Hawkes process
  modeling of {COVID}-19 with mobility leading indicators and spatial
  covariates},'' \emph{\BIBforeignlanguage{en}{International Journal of
  Forecasting}}, vol.~38, no.~2, pp. 505--520, 2022.

\bibitem{Schoenberg2023}
F.~Schoenberg, ``\BIBforeignlanguage{en}{Estimating {Covid}-19 transmission
  time using {Hawkes} point processes},'' \emph{\BIBforeignlanguage{en}{The
  Annals of Applied Statistics}}, vol.~17, no.~4, 2023.

\bibitem{Sun2006}
J.~Sun, \emph{The statistical analysis of interval-censored failure time
  data}.\hskip 1em plus 0.5em minus 0.4em\relax Springer, 2006, vol.~3, no.~1.

\bibitem{Chen2012}
D.-G.~D. Chen, J.~Sun, and K.~E. Peace, \emph{Interval-censored time-to-event
  data: methods and applications}.\hskip 1em plus 0.5em minus 0.4em\relax CRC
  Press, 2012.

\bibitem{Bogaerts2017}
K.~Bogaerts, A.~Komarek, and E.~Lesaffre, \emph{Survival analysis with
  interval-censored data: a practical approach with examples in R, SAS, and
  BUGS}.\hskip 1em plus 0.5em minus 0.4em\relax Chapman and Hall/CRC, 2017.

\bibitem{Du2021}
M.~Du and J.~Sun, ``Statistical analysis of interval-censored failure time
  data,'' \emph{Chinese Journal of Applied Probability and Statistics},
  vol.~37, no.~6, pp. 627--654, 2021.

\bibitem{Chen2023}
L.-P. Chen and B.~Qiu, ``Analysis of length-biased and partly interval-censored
  survival data with mismeasured covariates,'' \emph{Biometrics}, vol.~79,
  no.~4, pp. 3929--3940, 2023.

\bibitem{Ma2014}
L.~Ma, Y.~Feng, D.-G.~D. Chen, and J.~Sun, ``Interval-censored time-to-event
  data and their applications in clinical trials,'' \emph{Clinical Trial
  Biostatistics and Biopharmaceutical Applications}, vol. 307, 2014.

\bibitem{Kong2023}
Q.~Kong, P.~Calderon, R.~Ram, O.~Boichak, and M.-A. Rizoiu, ``Interval-censored
  transformer hawkes: Detecting information operations using the reaction of
  social systems,'' in \emph{ACM Web Conference}, 2023.

\bibitem{Shen2014}
H.~Shen, D.~Wang, C.~Song, and A.-L. Barab{\'a}si, ``Modeling and predicting
  popularity dynamics via reinforced poisson processes,'' in \emph{AAAI},
  vol.~28, no.~1, 2014.

\bibitem{Ogata1981}
Y.~Ogata, ``\BIBforeignlanguage{en}{On {Lewis}' simulation method for point
  processes},'' \emph{\BIBforeignlanguage{en}{IEEE Transactions on Information
  Theory}}, vol.~27, no.~1, pp. 23--31, 1981.

\bibitem{dassios2013exact}
A.~Dassios and H.~Zhao, ``Exact simulation of hawkes process with exponentially
  decaying intensity,'' 2013.

\bibitem{rizoiu2018sir}
M.-A. Rizoiu, S.~Mishra, Q.~Kong, M.~Carman, and L.~Xie, ``Sir-hawkes: Linking
  epidemic models and hawkes processes to model diffusions in finite
  populations,'' in \emph{Proceedings of the 2018 world wide web conference},
  2018, pp. 419--428.

\bibitem{appendix}
``Appendix: Linking across data granularity: Fitting multivariate hawkes
  processes to partially interval-censored data,'' 2024,
  \url{https://bit.ly/3UAKpt6}.

\bibitem{Phillips2003}
C.~L. Phillips, J.~M. Parr, E.~A. Riskin, and T.~Prabhakar, \emph{Signals,
  systems, and transforms}.\hskip 1em plus 0.5em minus 0.4em\relax Prentice
  Hall Upper Saddle River, 2003.

\bibitem{Ogata1978}
Y.~Ogata, ``\BIBforeignlanguage{en}{The asymptotic behaviour of maximum
  likelihood estimators for stationary point processes},''
  \emph{\BIBforeignlanguage{en}{Annals of the Institute of Statistical
  Mathematics}}, vol.~30, no.~2, pp. 243--261, Dec. 1978.

\bibitem{Wachter2006}
A.~Wächter and L.~T. Biegler, ``\BIBforeignlanguage{en}{On the implementation
  of an interior-point filter line-search algorithm for large-scale nonlinear
  programming},'' \emph{\BIBforeignlanguage{en}{Mathematical Programming}},
  vol. 106, no.~1, pp. 25--57, Mar. 2006.

\bibitem{Fokianos2009}
K.~Fokianos, A.~Rahbek, and D.~Tj{\o}stheim, ``Poisson autoregression,''
  \emph{Journal of the American Statistical Association}, vol. 104, no. 488,
  pp. 1430--1439, 2009.

\bibitem{Eckner2012}
E.~Erdogan, S.~Ma, A.~Beygelzimer, and I.~Rish, \emph{Statistical Models for
  Unequally Spaced Time Series}, pp. 626--630.

\bibitem{Rehfeld2011}
K.~Rehfeld, N.~Marwan, J.~Heitzig, and J.~Kurths, ``Comparison of correlation
  analysis techniques for irregularly sampled time series,'' \emph{Nonlinear
  Processes in Geophysics}, vol.~18, no.~3, pp. 389--404, 2011.

\bibitem{Wu2019}
S.~Wu, M.-A. Rizoiu, and L.~Xie, ``{Estimating Attention Flow in Online Video
  Networks},'' \emph{ACM HCI}, vol.~3, no. CSCW, pp. 1--25, nov 2019.

\bibitem{Dong2020}
E.~Dong, H.~Du, and L.~Gardner, ``\BIBforeignlanguage{en}{An interactive
  web-based dashboard to track {COVID}-19 in real time},''
  \emph{\BIBforeignlanguage{en}{The Lancet Infectious Diseases}}, vol.~20,
  no.~5, pp. 533--534, May 2020.

\bibitem{Aylien2020}
Aylien, ``{AYLIEN} {Coronavirus} {Dataset},'' 2020.

\bibitem{vdMaaten2008}
L.~van~der Maaten and G.~Hinton, ``Visualizing data using t-sne,''
  \emph{Journal of Machine Learning Research}, vol.~9, no.~86, pp. 2579--2605,
  2008.

\bibitem{lemonnier2017multivariate}
R.~Lemonnier, K.~Scaman, and A.~Kalogeratos, ``Multivariate hawkes processes
  for large-scale inference,'' in \emph{Proceedings of the AAAI conference on
  artificial intelligence}, vol.~31, no.~1, 2017.

\bibitem{nickel2020learning}
M.~Nickel and M.~Le, ``Modeling sparse information diffusion at scale via lazy
  multivariate hawkes processes,'' in \emph{Proceedings of the Web Conference
  2021}, 2021, pp. 706--717.

\end{thebibliography}

% \vspace*{-2\baselineskip}
% \begin{IEEEbiography}[{\includegraphics[width=1in,height=1.25in,clip,keepaspectratio]{bio/pio.png}}]{Pio Calderon}
% is a PhD candidate at the University of Technology Sydney, where his research primarily explores self-exciting models of information spread within online environments. He received his Bachelor’s degree in Physics and Master’s degree in Applied Mathematics from the University of the Philippines Diliman.
% \end{IEEEbiography}

% \vspace*{-2\baselineskip}
% \begin{IEEEbiography}[{\includegraphics[width=1in,height=1.25in,clip,keepaspectratio]{bio/alex.png}}]{Alexander Soen}
% received a Bachelor of Advanced Computing (R\&D) with First Class Honours from the Australian National University (ANU), Australia, in 2019. 
% He is currently pursuing a PhD at ANU, with research interests focused on machine learning, particularly in boosting algorithms, information geometric tools, and the theory of loss functions.
% \end{IEEEbiography}
    
% \vspace*{-2\baselineskip}
% \begin{IEEEbiography}[{\includegraphics[width=1in,height=1.25in,clip,keepaspectratio]{bio/andrei.png}}]{Marian-Andrei Rizoiu}
% is an Associate Professor leading the Behavioral Data Science lab at the University of Technology Sydney. 
% His interdisciplinary research crosses computer and social sciences, blending psycholinguistics, digital communication and stochastic modelling to understand human attention dynamics in the online environment, the emergence of influence and opinion polarization.
% \end{IEEEbiography}

\clearpage
\includepdf[pages=-]{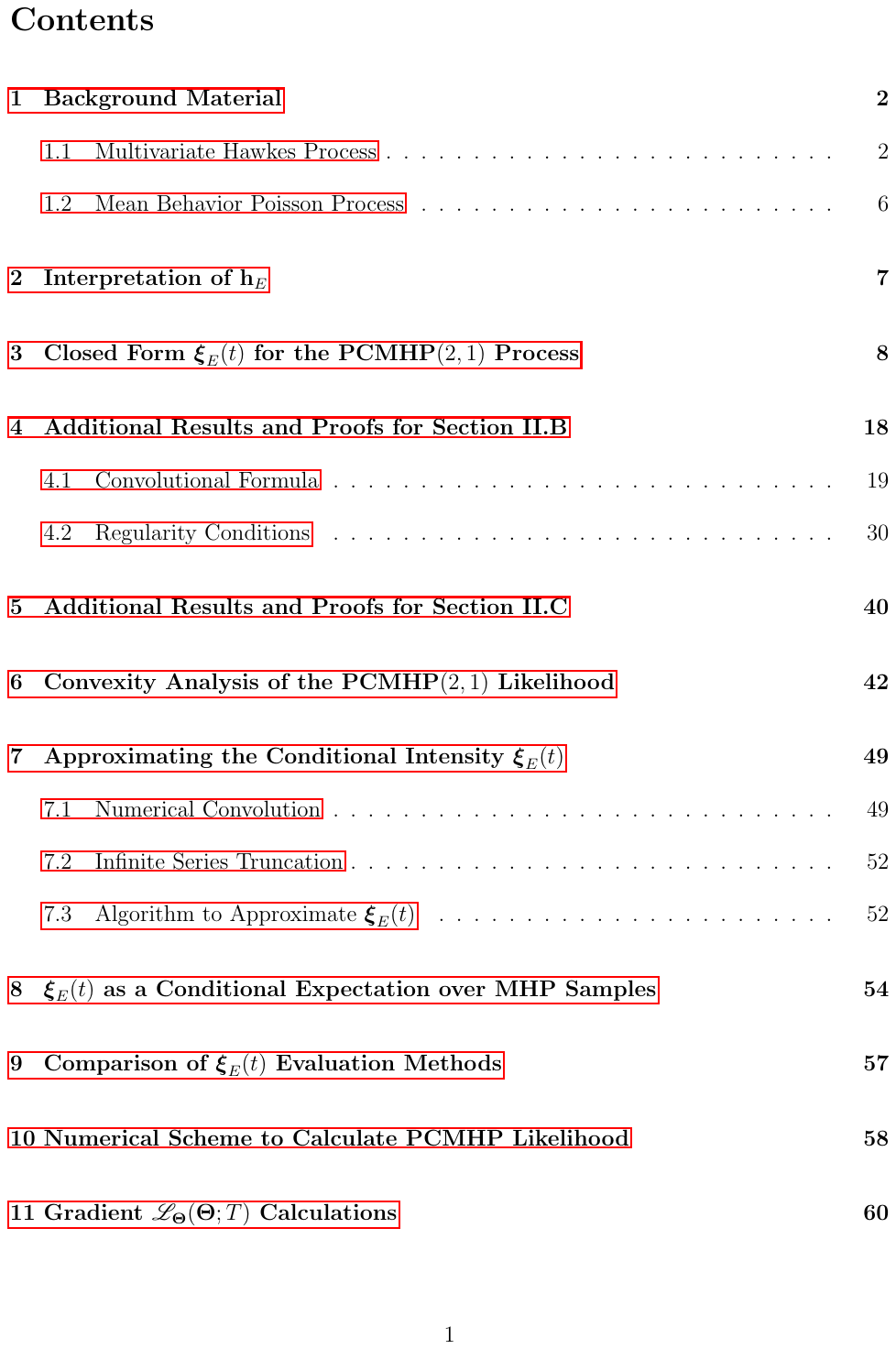}

\end{document}